\documentclass{article}

\usepackage{arxiv}

\usepackage[utf8]{inputenc} 
\usepackage[T1]{fontenc}    
\usepackage{hyperref}       
\usepackage{url}            
\usepackage{booktabs}       
\usepackage{amsfonts}       
\usepackage{nicefrac}       
\usepackage{microtype}      
\usepackage{lipsum}
\usepackage{graphicx}
\graphicspath{ {./images/} }
\usepackage{times}
\usepackage{soul}
\usepackage{url}
\usepackage[utf8]{inputenc}
\usepackage[small]{caption}
\usepackage{graphicx}
\usepackage{xcolor}
\usepackage{amsmath}
\usepackage{amsthm}
\usepackage{amsfonts}
\usepackage{subcaption}
\usepackage{booktabs}
\usepackage{algorithm}
\usepackage{algorithmic}
\usepackage[switch]{lineno}
\usepackage{multirow}

\usepackage{soul}

\title{Reasoning Capabilities of Large Language Models. \\ Lessons Learned from General Game Playing}

\author{
  Maciej {\'S}wiechowski \\
  Grail Team \\
  Warsaw, Poland\\
  \texttt{maciej@grailteam.com} \\
   \And
  Adam {\.Z}ychowski \\
  Warsaw University of Technology \\
  Warsaw, Poland\\
  \texttt{a.zychowski@mini.pw.edu.pl} \\
   \And
  Jacek Ma{\'n}dziuk \\
  Warsaw University of Technology \\
  Warsaw, Poland\\
  AGH University of Krakow \\
  Cracow, Poland \\
  \texttt{mandziuk@mini.pw.edu.pl} \\
}

\begin{document}
\maketitle
\begin{abstract}
%
%
%
This paper examines the reasoning capabilities of Large Language Models (LLMs) from a novel perspective, focusing on their ability to operate within formally specified, rule-governed environments. We evaluate four LLMs (\textit{Gemini 2.5} Pro and Flash variants, \textit{Llama 3.3 70B} and \textit{GPT-OSS 120B}) on a suite of forward-simulation tasks—including next / multistep state formulation, and legal action generation—across a diverse set of reasoning problems illustrated through General Game Playing (GGP) game instances.
Beyond reporting instance-level performance, we characterize games based on 40 structural features and analyze correlations between these features and LLM performance.
Furthermore, we investigate the effects of various game obfuscations to assess the role of linguistic semantics in game definitions and the impact of potential prior exposure of LLMs to specific games during training.
The main results indicate that three of the evaluated models generally perform well across most experimental settings, with performance degradation observed as the evaluation horizon increases (i.e., with a higher number of game steps). Detailed case-based analysis of the LLM performance provides novel insights into common reasoning errors in the considered logic-based problem formulation, including hallucinated rules, redundant state facts, or syntactic errors.
%
Overall, the paper reports clear progress in formal reasoning capabilities of contemporary models.
\end{abstract}


\section{Introduction}
\label{sec:introduction}

While early applications of Large Language Models (LLMs) focused primarily on natural language tasks such as chatbots, translation, and text correction~\cite{wang2025history}, recent advances have significantly expanded their capabilities~\cite{webb2023emergent}. Modern LLMs are now frequently employed for tasks that require non-trivial reasoning, structured information processing, and multistep inference. As a result, they are increasingly relied upon in decision-support scenarios, knowledge-intensive workflows, and systems where incorrect reasoning may lead to costly or even critical errors~\cite{bo2025rely}. This growing reliance raises an important research question: to what extent can LLMs reliably perform reasoning 
in structured 
environments? Understanding both the capabilities and limitations of LLMs in such contexts is essential for building trustworthy LLM-based systems. 

In this work, we investigate this question using the General Game Playing (GGP) framework~\cite{genesereth2005general,swiechowskisurvey2015} as an experimental environment. GGP was originally introduced as a testbed for developing agents capable of operating across multiple games without prior domain-specific knowledge. Each game 
represents an abstract, well-defined problem expressed using a formal logic-based language known as the Game Description Language (GDL). The formal and declarative nature of GDL makes it particularly suitable for studying reasoning processes. This setting provides a unique opportunity to analyze the relationship between structural properties of reasoning problems and the performance of LLMs. In particular, we focus on the ability of LLMs to correctly simulate GGP games, i.e., to follow and apply logical rules, infer valid state transitions, and produce correct outcomes under controlled experimental conditions. Importantly, correctness can be evaluated objectively using a GDL simulator, which serves as a reliable ground truth. 

The motivation for this study stems from the growing tendency of users to delegate reasoning tasks to LLMs. In such scenarios, reasoning often proceeds iteratively: known information (facts) are combined with instructions (rules) to derive new facts, which then enable further inference steps. This form of symbolic reasoning resembles the mechanics of GGP/GDL, making the framework well suited for evaluating LLM reasoning under controlled and interpretable conditions.

The main contributions of this paper are five-fold:
\begin{enumerate}
     \item 
     An adaptation of the GGP framework as a benchmark for evaluating reasoning capabilities. 
     \item 
     A thorough evaluation and performance comparison of four contemporary models in the proposed reasoning setting. 
     \item 
     An analysis of how problem structure and compositional complexity correlate with reasoning accuracy.
    \item 
    An investigation of the semantic grounding by comparing performance on problems with meaningful (original) names and syntactically obfuscated representations. 
    \item 
    Insights presentation on common LLM mistakes in the proposed logic-based reasoning benchmark.
\end{enumerate}


\section{Related Work}
\label{sec:related-work}

\paragraph{Benchmarks for Logical Reasoning in LLMs.}
The rapid progress of LLMs has sparked interest in their ability to perform reasoning beyond surface-level pattern matching~\cite{kojima2022large}. Several benchmarks have been proposed to evaluate logical and multistep reasoning in LLMs. A recent survey~\cite{liu2025logical} provides a comprehensive overview of this area, highlighting that while reasoning capabilities have noticeably improved since around 2023, existing evaluation methodologies remain insufficient,
motivating the need for more rigorous and interpretable benchmarks.

Existing benchmark datasets include ProofWriter~\cite{tafjord2021proofwriter}, which evaluates deductive reasoning over natural language rules and facts, LogiQA~\cite{liu2021logiqa}, which focuses on logical reading comprehension, and LogicBench~\cite{parmar2024logicbench}, which systematically tests specific inference patterns using carefully constructed question-answer (QA) tasks. 
Following the intuition reported in~\cite{parmar2024logicbench}, longer rules are more challenging to handle, and larger models tend to be better logical reasoners.
While these benchmarks have been instrumental in advancing the field, they typically rely on relatively short reasoning chains, operate in a single-step question-answering format, and evaluate correctness primarily via final answer accuracy. In contrast, our work employs game testbeds, where reasoning unfolds across multiple state transitions, requires consistent variable bindings, and must respect a formally defined symbolic world model.

Beyond accuracy-based evaluation,~\cite{liualigning} propose a framework for measuring logical preference consistency, showing that similarity-based metrics such as BERTScore correlate better with human judgments than LLM-based evaluators. Similarly,~\cite{xu2025large} conduct a fine-grained analysis of reasoning errors across deductive, inductive, and abductive tasks, identifying common failure modes such as hallucination, incorrect process execution, and redundant explanations. 


\paragraph{Multistep Reasoning Limitations.}
A recurring conclusion across prior work is that LLMs perform reasonably well on single-step inference but struggle as compositional depth increases.~\cite{creswellselection} demonstrate that accuracy drops 
on multistep reasoning tasks even in relatively simple domains, such as bAbI~\cite{weston2015towards}, multi-hop QA~\cite{welbl2018constructing}, and ProofWriter~\cite{tafjord2021proofwriter}. 
In the latter, the proposed Selection-Inference framework mitigates some of these issues by decomposing reasoning into explicit stages.

Other studies attempt to augment LLMs with external symbolic components. For instance,~\cite{pan2023logic} integrate
symbolic solvers into the inference loop, showing substantial performance gains compared to standalone LLMs. 
However, such approaches blur the boundary between the reasoning capability of the language model itself and that of the auxiliary solver. In contrast, our work intentionally evaluates LLMs as standalone approximate reasoners and compares their outputs against an exact symbolic interpreter.

In~\cite{wang2024can},
the authors 
test LLMs using automatically generated inferential rules of varying complexity, revealing particular difficulties with compositional and disjunctive-transitive rules. Notably, the performance degrades as reasoning (compositional) depth increases, a trend also observed for human participants. However, the benchmark is based on single-step patterns formulated as if-then statements,
which is arguably less complex than GDL games.

~\cite{sullivan2025can} state that whether LLMs can truly act as symbolic reasoners remains an open question, warranting controlled experimental setups that disentangle symbolic structure from linguistic familiarity -- an issue we address explicitly through symbol obfuscation experiments.

\paragraph{Games as Reasoning Evaluation Environments.}
In~\cite{schultzmastering}, the authors argue that games expose LLMs’ difficulty in consistently reasoning over future states and propose external planning mechanisms to mitigate this limitation. Their evaluation uses ELO scores in four games.
Our work, however, targets a more fundamental problem of the correctness of forward simulation itself, independent of strategic optimality, and across a substantially broader set of games.

Several studies evaluate LLMs as game-playing agents.
Early analyses already noted that models such as GPT-3 generate fluent responses but exhibit systematic reasoning failures in interactive, rule-governed settings~\cite{sobieszek2022playing}. Subsequent empirical benchmarks confirm these limitations.~\cite{fan2024can} show that LLMs struggle with belief refinement and uncertainty in game-theoretic scenarios, while~\cite{hu2025lmgame} 
report poor long-horizon performance and strong sensitivity to prompt design across a diverse suite of games.
Competitive evaluation frameworks further reveal persistent issues such as performing invalid actions 
in multiple games~\cite{topsakal2024evaluating}. 

A common limitation of these approaches is that formal game rules are either implicit or translated into natural language, which conflates logical inference with language understanding and prompt adherence. Moreover, many benchmarks rely on relative performance (e.g., win rates or head-to-head comparisons) rather than absolute correctness with respect to the game 
formalism. 

The work particularly related to 
our setting is~\cite{li2025gvgai}, 
based on the Video Game Description Language (VGDL), which is conceptually similar to GDL. However, they use VGDL primarily to enable benchmark construction, and 
their evaluation focuses on playing efficacy rather than reasoning over the game description itself. 

Our work is also related to studies on internal world models induced by sequence prediction.~\cite{yuan2025revisiting} revisit the Othello World Model Hypothesis, showing that models trained on move sequences can internalize latent board representations. While this demonstrates that LLMs can acquire implicit world models through extensive task-specific training, our experiments instead probe whether 
LLMs can correctly simulate symbolic dynamics across multiple games without such retraining, thereby assessing generalization rather than memorization.
Finally, in~\cite{tanaka2024grammar}, the authors explore the inverse problem of generating GDL-like game descriptions using LLMs, reporting coherent but imperfect outputs with increasing errors for longer descriptions. They propose external iterative improvement through grammar guidance. 

\section{GDL Background}

\label{sec:ggp-gdl}

We adopt the 
GGP framework 
originally proposed by the Stanford Logic Group~\cite{genesereth2005general}. Instead of relying on hard-coded domain knowledge, GGP systems receive the rules of a game at runtime, expressed in GDL~\cite{thielscher2011general}.

GDL is a first-order logic programming language based on Datalog~\cite{greco2015datalog}, which is a subset of Prolog. 
In this work, we use the prefix KIF notation which was the standard representation used in the official GGP competitions. 
A GDL game description is provided as a text file containing a set of facts and rules (the ordering is irrelevant)
that specify the complete logical structure of a game. This enables automated systems to infer legal moves, state transitions, and terminal conditions based solely on logical reasoning.

\paragraph{Facts.}
Facts are declarative statements that serve as building blocks of state representation. Syntactically, facts are ground predicates, i.e., logical clauses without variables, which are assumed to be \emph{true} within a given context.
Facts are enclosed in parentheses (and can be nested). They consists of a predicate name (representing the type of relation) followed by a sequence of arguments (attributes). 
For example, the fact \texttt{(location pacman 5 3)} states that the entity \emph{pacman} is at coordinates $(5,3)$.

Facts may be static (constant) or dynamic (state-dependent), which is distinguished using GDL keywords: \emph{init}, \emph{true}, and \emph{next}, which are discussed 
below.

\paragraph{Rules.}
%
Rules constitute the core mechanism for expressing logic in GDL. They define how new facts can be derived from existing ones. Each rule represents a logical implication and consists of a \emph{head} (the consequence) and a \emph{body} (the conditions). The implication operator \texttt{<=} separates these two parts.

The \emph{head} is a single term that can be viewed as a template for facts that the rule produces if it is satisfied. The head can either be a ground fact or a fact-like term containing variables. In the latter case, each valid grounding of 
variables with concrete symbols results in a distinct derived fact. Variables 
are denoted by identifiers starting with the character \texttt{?}. During the resolution process, variables with the same name must be consistently substituted with the same symbols throughout the entire rule.

The rule body consists of conditions (either positive or negated). A rule is considered satisfied if all conditions evaluate to \emph{true} under a consistent variable assignment. A condition 
is satisfied if at least one matching fact can be derived from the current state, either directly through \emph{init} or \emph{next} facts, or indirectly through the application of other rules. 

A simple rule in Tic-Tac-Toe may look as follows:
\begin{verbatim}
(<= (column ?n ?x)
    (true (cell 1 ?n ?x))
    (true (cell 2 ?n ?x))
    (true (cell 3 ?n ?x)))
\end{verbatim}

This rule states that if the symbol, which substitutes variable \texttt{?x}, appears in all three rows of a given column \texttt{?n}, then the fact \texttt{(column ?n ?x)} holds. Consequently, for each column in which this condition is satisfied, a corresponding fact is derived, such as \texttt{(column 2 x)} or \texttt{(column 1 o)}. Another rule may use facts of the type ``column'' as arguments. Overall, reasoning in GDL involves multi-branch logical inference with consistent variable bindings, which poses challenges for LLMs due to non-local dependencies.\\

GDL follows a closed-world assumption (CWA): any fact that cannot be derived (either explicitly or by at least one rule) is considered to be \emph{false}.

Apart from a small set of reserved keywords (discussed below), the choice of predicate and symbol names is left entirely to the designer of the game description. Two descriptions that use different symbols may represent the same problem (game), as long as the syntactic dependencies are preserved.

\paragraph{State Transition and Selected Keywords.}
%
The current state of a game is modeled through explicit state transitions defined using the dedicated keywords \emph{init} and \emph{next}. Each expression of the form $(init\ f)$ specifies that a fact $f$ holds in the initial state.

Player names are introduced using the keyword $(role\ r)$, which declares $r$ as a game participant and action-maker.

The predicate $(legal\ r\ a)$ specifies that role $r$ is allowed to perform action $a$ in the current state. Legal actions are derived by evaluating rules over the current set of facts and determine the available choices for each player.

Once an action is selected, the simulator injects facts of the form $(does\ r\ a)$, indicating that role $r$ has executed action $a$ in the current state. These facts represent the joint move of all players and serve as inputs to the state transition rules.

State transitions are defined by rules with consequences expressed using the head $(next\ t)$. When such a rule is satisfied, each resulting ground term $t$ is produced as a fact in the subsequent state. For example, the following \emph{next} rule specifies that a cell \texttt{?u ?v} becomes empty (represented by the blank symbol \texttt{b}) if a player moves a piece \texttt{?p} from that cell:
\begin{verbatim}
(<= (next (cell ?u ?v b))
    (does ?player (move ?p ?u ?v ?x ?y)))
\end{verbatim}

After each transition, the previous state is discarded and replaced by the newly derived set of facts (all at once) in an atomic operation, analogous to a transaction in a database system. Consequently, all dynamic facts must be explicitly re-derived at every step based on the previous state, i.e., any fact that is not derived through a \emph{next} rule is assumed to be \emph{false} in the subsequent state.


\section{Experimental Setup}
\label{sec:setup}

\paragraph{Experimental Tasks.}
To assess LLMs' symbolic reasoning within GGP, we designed four tasks of various nature and complexity, 
including one step / multistep game simulations, 
rule interpretation, precondition verification, and state consistency.
\begin{itemize}
    \item \textbf{Task 1: Next state generation:} Given a state and a joint move (one action per player), 
    provide the resulting state. Tests reasoning with \emph{next} rules and state persistence.
     \item \textbf{Task 2: Legal actions generation:} Given a state, enumerate all actions currently available to the players. Evaluates reasoning with \emph{legal} rules and correct traversal of rule dependencies.
    \item \textbf{Task 3: Multistep state generation:} From the initial state of a game, given a sequence of $n$ joint moves, provide the resulting state. Requires reasoning with \emph{init} and \emph{next} rules and implicit tracking of intermediate states.
       \item \textbf{Task 4: Multistep action-state generation:} From the initial state of a game, generate a valid sequence of $n$ consecutive actions and the resulting state. Tests integrated reasoning with \emph{init}, \emph{legal}, and \emph{next} rules while maintaining consistency over multiple steps.
\end{itemize}

\paragraph{Semantic Grounding and Obfuscation.}
A critical research question is whether LLMs perform symbolic reasoning or rely on linguistic correlations associated with meaningful identifiers (e.g., predicting that an entity \texttt{rook} moves horizontally based on the pre-training chess data rather than GDL rules). To investigate this, all four tasks are conducted in two variants:
\begin{enumerate}
    \item \textbf{Original:} Using human-readable GDL descriptions with meaningful names (e.g., \texttt{cell}, \texttt{capture}, etc.).
    \item \textbf{Obfuscated:} Using structurally identical descriptions where all non-keyword predicates and constants are replaced according to one of the three following variants: (i) \textit{placeholder terms} - replacing symbols with generic identifiers: \texttt{term1}, \texttt{term2}, ...; (ii) \textit{dictionary words} - replacing symbols with semantically neutral nouns, e.g., ``chair'', ``forest'', etc.; or (iii) \textit{random strings} - replacing symbols with randomly generated strings consisting of 5 to 8 alphanumeric characters.
\end{enumerate}
Comparing performance across these variants allows us to isolate the model's pure symbolic reasoning capability from its semantic knowledge base.

\paragraph{Game Instances.}
%
Initially, we downloaded a pool of 86 GDL game descriptions 
from public repositories~\cite{stanford-gdl,dresden-gdl}, from which we selected a representative subset of 35 games for the main experiments. The selection was based on preliminary single-step transition prediction experiments and diversity of structural properties of the games. Definitions of all considered games are provided in the supplementary material.

\paragraph{Evaluated Models.}
To ensure the generalizability of our findings and to analyze the impact of model size and specificity (e.g., Gemini Pro. vs Flash) on symbolic reasoning, we selected four LLMs representing different points in the performance-efficiency spectrum: 
two proprietary state-of-the-art models of \textit{Gemini 2.5} family (Pro and Flash variants)~\cite{comanici2025gemini} and two popular open-weights models (\textit{Llama 3.3 70B}~\cite{dubey2024llama} and \textit{GPT-OSS 120B}~\cite{agarwal2025gpt}).
This diverse set of models allows us to check 
the 
impact of model architecture (open vs. closed) and minimize the influence of a single model family's biases.

The prompting methodology is described in the 
supplementary material.

\paragraph{Performance Metrics.}
We employ two performance 
metrics tailored for the reasoning problem. As LLM outputs are expected to be sets of facts following the GDL formalism, the first metric is based on the Jaccard index:
\begin{equation}
JI(Y,Y') = \frac{1}{N} \sum_{i=1}^{N} \frac{|Y_i \cap Y'_i|}{|Y_i \cup Y'_i|}  
\end{equation}
where $i$ denotes the current test sample, $Y_i$ is the expected set of GDL facts, and $Y'_i$ is the set of facts in the model's output. Additionally, we keep track of the number of missing or superfluous facts for a detailed analysis.

Since a mistake may result in cascading the errors in subsequent forward simulation steps, and because in many games a large portion of the game state is rewritten at each transition, we also introduce 
another metric:
\begin{equation}
\%S = \frac{1}{N} \sum_{i=1}^{N} \mathbb{I}\!\left( JI(Y,Y') = 1 \right)
\end{equation}
where $\mathbb{I}(\cdot)$ is the indicator function.

If 
an LLM output cannot be parsed into a valid set of GDL facts even under relaxed parsing rules, or if it contains illegal actions in generation tasks, the corresponding sample is assigned a performance score of $0$ in both metrics. 

\section{Results}
\label{sec:results}

The main results are presented in Tables~\ref{tab:next_state}-~\ref{tab:generation} corresponding to Tasks 1- 4.
%
Due to space limits. each table 
shows individual results for $8$ games (reasoning problems) and average 
scores over all 35 games used in the experiments. These 8 games were selected 
based on the highest average number of facts in state representations, while keeping only one representative from each 
subset of 
similar games, e.g., \texttt{checkers} over \texttt{checkers-mustjump}. The results for all $35$ games, as well as 
other experimental settings (including 
other reasoning horizons $n$) are provided in the supplementary material. 

\textit{Generally, across all tasks, Gemini~2.5~Pro consistently achieves the highest average performance, followed by Gemini~2.5~Flash, GPT-OSS~120B, and Llama~3.3~70B. A Wilcoxon signed-rank test confirmed that this performance hierarchy is statistically significant across all tasks ($p < 0.05$), with the highest $p$-value of $0.003495$ recorded between the two Gemini models in the Next state generation task.}

The easiest task is Next state generation (Table~\ref{tab:next_state}), in which all but the weakest model achieve very high accuracy (Avg JI $>0.8$). Across all 35 games, Gemini~2.5~Pro generates fully correct successor states in over 95\% of the cases on average, and in 34 games it achieves $\%S \geq 0.85$. Given that several of the evaluated games have indeed high structural complexity (see complexity features in the supplementary material), these results indicate that contemporary LLMs are 
capable of accurately applying one step symbolic rule-based reasoning in a wide range of 
problems, representing a substantial improvement over conclusions reported in earlier literature~\cite{creswellselection,li2025gvgai}.

\begin{table}[ht]
\centering
\caption{
Next state generation results for selected games.}
\label{tab:next_state}
\setlength{\tabcolsep}{3pt}
\begin{tabular}{lcccccccc}
\toprule
 & \multicolumn{2}{c}{Gemini 2.5 Pro} & \multicolumn{2}{c}{Gemini 2.5 Flash} & \multicolumn{2}{c}{GPT-OSS 120B} & \multicolumn{2}{c}{Llama 3.3 70B} \\
\cmidrule(lr){2-3} \cmidrule(lr){4-5} \cmidrule(lr){6-7} \cmidrule(lr){8-9}
Game & JI & \%S & JI & \%S & JI & \%S & JI & \%S \\
\midrule
battlebrushes & 0.997 & 0.950 & 0.997 & 0.950 & 0.988 & 0.750 & 0.945 & 0.100 \\
bomberman2p & 0.995 & 0.950 & 0.972 & 0.700 & 0.932 & 0.500 & 0.946 & 0.350 \\
checkers & 1.000 & 1.000 & 0.999 & 0.950 & 0.994 & 0.850 & 0.956 & 0.250 \\
chess & 0.998 & 0.850 & 0.998 & 0.850 & 0.996 & 0.800 & 0.973 & 0.150 \\
mummymaze2p & 1.000 & 1.000 & 0.999 & 0.900 & 1.000 & 1.000 & 0.989 & 0.300 \\
pacman3p & 1.000 & 1.000 & 0.994 & 0.750 & 1.000 & 1.000 & 0.989 & 0.650 \\
platformJumpers & 0.998 & 0.850 & 0.997 & 0.750 & 0.993 & 0.750 & 0.940 & 0.000 \\
snake\_2009\_big & 0.998 & 0.850 & 0.992 & 0.800 & 0.965 & 0.400 & 0.942 & 0.000 \\
\midrule
\textbf{Avg (all 35 games)} & \textbf{0.995} & \textbf{0.956} & 0.977 & 0.836 & 0.975 & 0.806 & 0.823 & 0.313 \\
\bottomrule
\end{tabular}
\end{table}

Task 2, Legal actions generation, is noticeably more difficult than Next state generation. While Jaccard Index values remain relatively high for strong models, strict success rates (\%S) visibly drops.
%
In particular, Gemini~2.5~Pro achieves $\%S < 0.5$ in two of the eight selected games (\texttt{chess} and \texttt{platformJumpers}), both of which have large and highly conditional action spaces, as well as in six out of the total 35 games. This gap between $JI$ and $\%S$ suggests that models often identify quite many correct actions but fail to produce a complete and exact set of legal moves.

A comparison of Task 1 and Task 2 results indicates that models are better at forward application of known actions than at discovering all possible actions. 
This difference could be attributed to the different conceptual structure of these tasks. A state, although represented as a set of facts, forms a single coherent concept, whereas each legal action constitutes a separate concept.
\begin{table}[ht]
\centering
\caption{Legal 
actions generation results for selected games.}
\label{tab:legal_moves}
\setlength{\tabcolsep}{3pt}
\begin{tabular}{lcccccccc}
\toprule
 & \multicolumn{2}{c}{Gemini 2.5 Pro} & \multicolumn{2}{c}{Gemini 2.5 Flash} & \multicolumn{2}{c}{GPT-OSS 120B} & \multicolumn{2}{c}{Llama 3.3 70B} \\
\cmidrule(lr){2-3} \cmidrule(lr){4-5} \cmidrule(lr){6-7} \cmidrule(lr){8-9}
Game & JI & \%S & JI & \%S & JI & \%S & JI & \%S \\
\midrule
battlebrushes & 1.000 & 1.000 & 1.000 & 1.000 & 1.000 & 1.000 & 0.536 & 0.000 \\
bomberman2p & 1.000 & 1.000 & 0.928 & 0.500 & 0.965 & 0.850 & 0.525 & 0.000 \\
checkers & 0.954 & 0.800 & 0.813 & 0.300 & 0.201 & 0.050 & 0.004 & 0.000 \\
chess & 0.783 & 0.150 & 0.311 & 0.000 & 0.371 & 0.000 & 0.143 & 0.000 \\
mummymaze2p & 1.000 & 1.000 & 1.000 & 1.000 & 0.967 & 0.800 & 0.910 & 0.000 \\
pacman3p & 1.000 & 1.000 & 0.954 & 0.750 & 0.982 & 0.900 & 0.483 & 0.000 \\
platformJumpers & 0.873 & 0.100 & 0.805 & 0.350 & 0.937 & 0.800 & 0.147 & 0.000 \\
snake\_2009\_big & 1.000 & 1.000 & 0.950 & 0.950 & 0.975 & 0.950 & 0.413 & 0.000 \\
\midrule
\textbf{Avg (all 35 games)} & \textbf{0.963} & \textbf{0.839} & 0.886 & 0.691 & 0.865 & 0.711 & 0.486 & 0.141 \\
\bottomrule
\end{tabular}
\end{table}

The Multistep 
state generation task 
(Table~\ref{tab:prediction}) is substantially more challenging than Next step 
generation, as it requires maintaining symbolic consistency across multiple successive transitions. While Gemini~2.5~Pro achieves the highest overall performance (Avg $JI = 0.865$, $\%S = 0.734$), it still fails to reach an exact final state in approximately one quarter of 
test sequences. Gemini~2.5~Flash attains a comparable Avg $JI = 0.843$ but a noticeably lower strict success rate of $\%S = 0.591$, indicating that small inconsistencies frequently accumulate even when most state facts are predicted correctly. The remaining models degrade more sharply, especially Llama~3.3~70B.

Performance also varies considerably across games. For example, Gemini~2.5~Pro achieves perfect or near-perfect results in \texttt{mummymaze2p} and \texttt{pacman3p} ($\%S = 1.0$ and $0.95$, respectively), but drops to $\%S = 0.30$ in \texttt{battlebrushes}, where complex interactions between agents and environment features amplify early errors. 
%
\begin{table}[ht]
\centering
\caption{Multistep state generation
results for selected games (n=5).}
\label{tab:prediction}
\setlength{\tabcolsep}{3pt}
\begin{tabular}{lcccccccc}
\toprule
 & \multicolumn{2}{c}{Gemini 2.5 Pro} & \multicolumn{2}{c}{Gemini 2.5 Flash} & \multicolumn{2}{c}{GPT-OSS 120B} & \multicolumn{2}{c}{Llama 3.3 70B} \\
\cmidrule(lr){2-3} \cmidrule(lr){4-5} \cmidrule(lr){6-7} \cmidrule(lr){8-9}
Game & JI & \%S & JI & \%S & JI & \%S & JI & \%S \\
\midrule
battlebrushes & 0.300 & 0.300 & 1.000 & 1.000 & 0.197 & 0.000 & 0.427 & 0.000 \\
bomberman2p & 1.000 & 1.000 & 0.936 & 0.450 & 0.407 & 0.050 & 0.020 & 0.000 \\
checkers & 0.791 & 0.650 & 0.816 & 0.550 & 0.639 & 0.400 & 0.000 & 0.000 \\
chess & 0.993 & 0.650 & 0.921 & 0.200 & 0.663 & 0.050 & 0.134 & 0.000 \\
mummymaze2p & 1.000 & 1.000 & 0.865 & 0.600 & 0.780 & 0.050 & 0.964 & 0.000 \\
pacman3p & 0.999 & 0.950 & 0.924 & 0.800 & 0.371 & 0.100 & 0.826 & 0.000 \\
platformJumpers & 0.848 & 0.800 & 0.793 & 0.400 & 0.618 & 0.150 & 0.000 & 0.000 \\
snake\_2009\_big & 0.894 & 0.500 & 0.897 & 0.650 & 0.828 & 0.050 & 0.855 & 0.000 \\
\midrule
\textbf{Avg (all 35 games)} & \textbf{0.865} & \textbf{0.734} & 0.843 & 0.591 & 0.688 & 0.389 & 0.493 & 0.133 \\
\bottomrule
\end{tabular}
\end{table}

The Multistep action--state generation task (Table~\ref{tab:generation}) is the most demanding setting, as it requires jointly selecting legal actions and correctly applying transition rules over multiple steps. Even for Gemini~2.5~Pro, performance drops relative to 
Task 3 (Avg $JI = 0.808$, $\%S = 0.653$), indicating that errors in action selection further compound state-tracking inaccuracies. Gemini~2.5~Flash degrades more sharply (Avg $JI = 0.616$, $\%S = 0.469$), followed by GPT-OSS~120B (Avg $JI = 0.561$, $\%S = 0.347$). Llama~3.3~70B performs poorly in this setting, suggesting that 
the model fails to reliably follow a  multistep decision process.


Overall, the results obtained by the LLMs for Tasks 1–4 suggest that their use in logic-based reasoning tasks should be approached with a certain caution, depending on the type of task and the required level of reliability. If result correctness at a level of at least 0.95 is required, this threshold is achieved only in Task 1 and only by the strongest model.
\begin{table}[ht]
\centering
\caption{
Multistep action-state generation results for selected games (n=5).}
\label{tab:generation}
\setlength{\tabcolsep}{3pt}
\begin{tabular}{lcccccccc}
\toprule
 & \multicolumn{2}{c}{Gemini 2.5 Pro} & \multicolumn{2}{c}{Gemini 2.5 Flash} & \multicolumn{2}{c}{GPT-OSS 120B} & \multicolumn{2}{c}{Llama 3.3 70B} \\
\cmidrule(lr){2-3} \cmidrule(lr){4-5} \cmidrule(lr){6-7} \cmidrule(lr){8-9}
Game & JI & \%S & JI & \%S & JI & \%S & JI & \%S \\
\midrule
battlebrushes & 0.500 & 0.500 & 0.868 & 0.850 & 0.213 & 0.000 & 0.087 & 0.000 \\
bomberman2p & 0.744 & 0.700 & 0.896 & 0.650 & 0.387 & 0.000 & 0.091 & 0.000 \\
checkers & 0.948 & 0.850 & 0.494 & 0.400 & 0.412 & 0.250 & 0.000 & 0.000 \\
chess & 0.915 & 0.100 & 0.763 & 0.250 & 0.633 & 0.150 & 0.000 & 0.000 \\
mummymaze2p & 0.994 & 0.950 & 0.961 & 0.750 & 0.796 & 0.000 & 0.145 & 0.000 \\
pacman3p & 0.898 & 0.850 & 0.400 & 0.400 & 0.310 & 0.050 & 0.000 & 0.000 \\
platformJumpers & 0.900 & 0.900 & 0.199 & 0.150 & 0.466 & 0.150 & 0.000 & 0.000 \\
snake\_2009\_big & 0.963 & 0.550 & 0.895 & 0.400 & 0.271 & 0.050 & 0.000 & 0.000 \\
\midrule
\textbf{Avg (all 35 games)} & \textbf{0.808} & \textbf{0.653} & 0.616 & 0.469 & 0.561 & 0.347 & 0.175 & 0.096 \\
\bottomrule
\end{tabular}
\end{table}

\subsection{Reasoning Horizon}
\label{sec:results-reasoning-horizon}

\begin{figure}[ht]
    \centering
    \begin{subfigure}[b]{0.47\columnwidth}
        \centering
        \includegraphics[width=\linewidth]{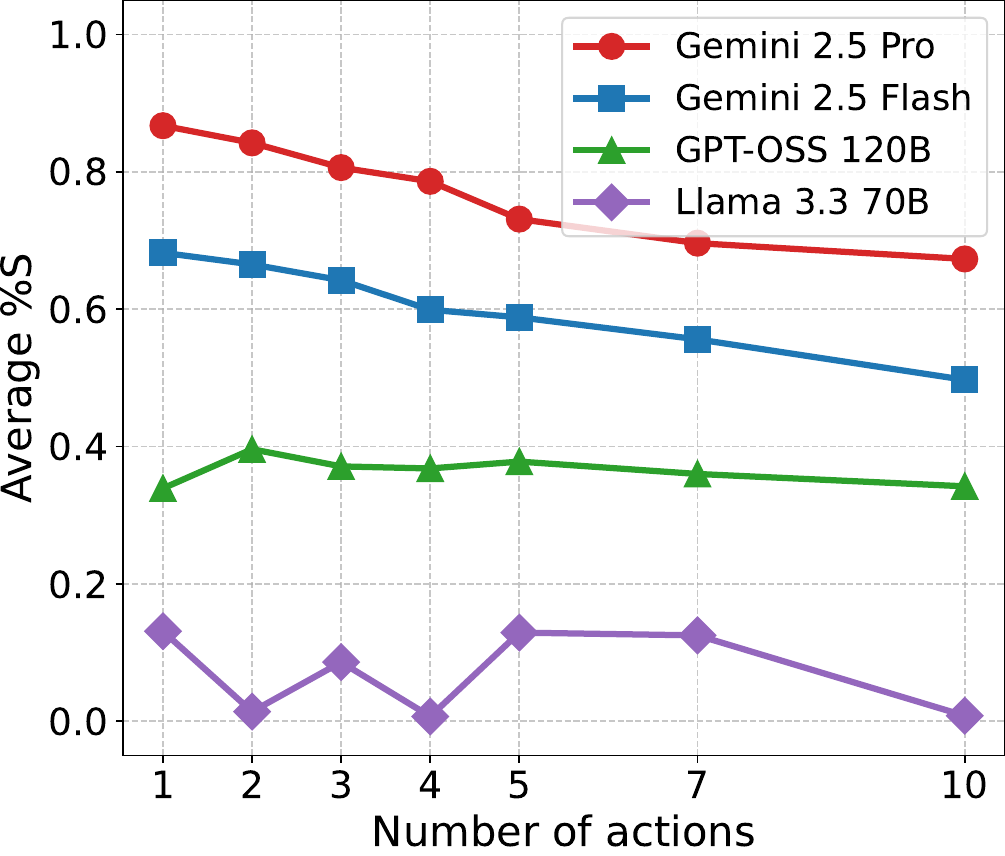}
        \caption{Original Games}
        \label{fig:normal}
    \end{subfigure}
    \hspace{1em}
    \begin{subfigure}[b]{0.47\columnwidth}
        \centering
        \includegraphics[width=\linewidth]{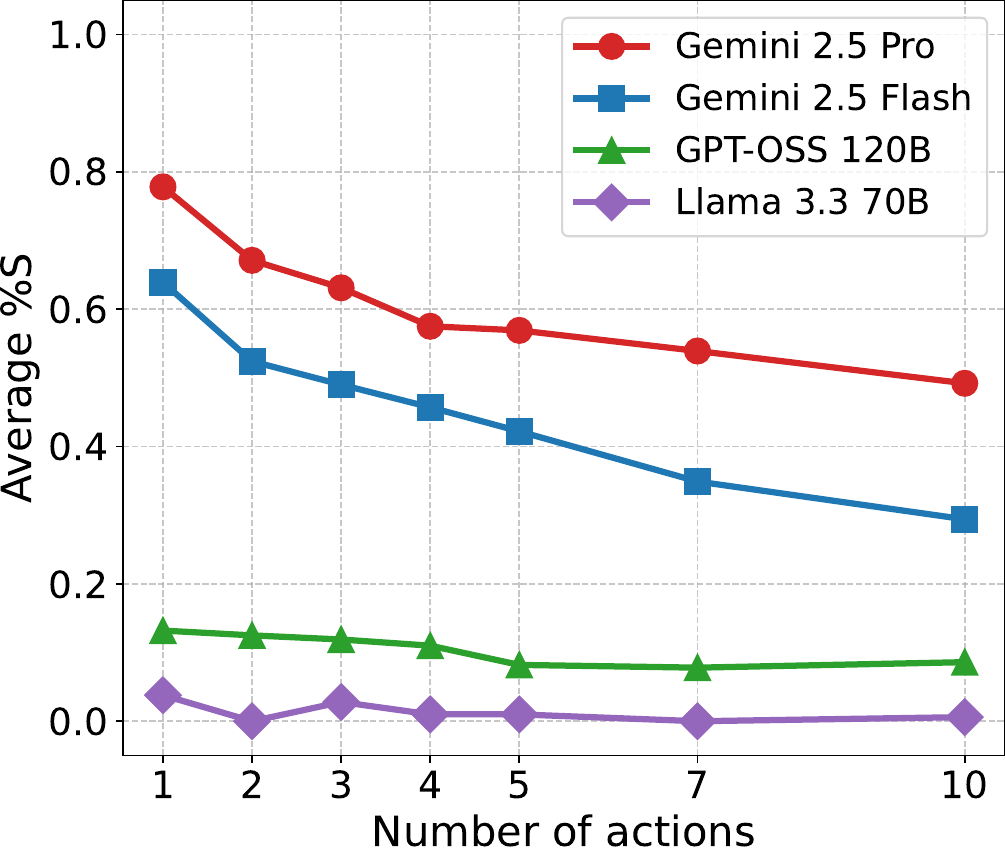}
        \caption{Obfuscated Games}
        \label{fig:obfuscated}
    \end{subfigure}
    \caption{Multistep state generation accuracy (\%S) across different horizon lengths. The left panel shows performance on original game descriptions, while the right panel presents results for the obfuscated `placeholder term' variant (detailed in Section~\ref{sec:obfuscation}).
    }
    \label{fig:multistep_results}
\end{figure}

To 
assess the 
impact of the reasoning horizon on the LLM Multistep state generation performance, we evaluate 
the models in Task~3
across varying sequence lengths. 
The results are summarized in Figure~\ref{fig:multistep_results}.
The performance hierarchy 
among the models remains consistent regardless of the prediction horizon. Gemini 2.5 Pro maintains a clear lead across all step counts, followed by Gemini 2.5 Flash, with the open-weights models (GPT-OSS and Llama 3.3) lagging behind. This suggests that the fundamental reasoning capability gap 
persists irrespective of task complexity. Notably, while the ranking is stable, the performance disparity between Gemini 2.5 Pro and Gemini 2.5 Flash becomes slightly more distinct, especially in the obfuscated variant, as the number of steps increases which likely stems from the 
Gemini 2.5 Pro superior ability to mitigate error propagation 
in longer deductive chains.

The degradation patterns differ between stronger and weaker reasoners. The high-performing models (Gemini family) exhibit a quasi-linear, monotonic decrease in accuracy ($\%S$) as the number of steps increases. In contrast, the weaker models, particularly Llama 3.3 70B, display significant volatility and irregularities. The non-monotonic fluctuations in the lower performance range shows that these models may often rely on spurious correlations or chance rather than a robust, step-by-step derivation of the state.

The comparison between the original (Figure~\ref{fig:normal}) and obfuscated (Figure~\ref{fig:obfuscated}) scenarios confirms the important role of semantic grounding. All models achieve lower scores on obfuscated games, but the impact of obfuscation is not the same across models. For the top two models, the performance gap between the semantic and obfuscated variants widens as the chain of reasoning lengthens. For instance, for Gemini 2.5 Pro the observed gap at one action is 0.089 and it increases to 0.181 at 10 actions. It may indicate that without the support of linguistic cues (e.g., meaningful predicate names like \texttt{capture} or \texttt{move}), the models struggle significantly more to maintain context and track symbol bindings over long horizons.

\subsection{Common Mistakes}
\label{sec:results-mistakes}

In this section, we focus common errors made by the two strongest models (Gemini 2.5 Pro and Gemini 2.5 Flash), as their failure modes are more indicative due to their overall higher performance.

\paragraph{Hallucinated or corrected rules.}
GDL descriptions are written by humans and may contain unintended omissions. In the game \texttt{farmers}, we identified two missing rules from a common-sense perspective: when a player buys or grows an item, previously held commodities of a different type are lost. After this discovery, we explicitly asked Gemini~2.5~Pro whether the GDL specification contained mistakes, and the model pointed out the absence of these rules. However, when reasoning solely based on the provided GDL description during the experiments, the models implicitly assumed that the missing constraints were present and generated state transitions accordingly. Since the game is playable despite these flaws, such silent ``corrections'' lead to formally incorrect outputs.

\paragraph{Extraneous facts.}
In games such as \texttt{checkersTiny} and \texttt{checkers-small}, the initial state contains facts describing both black and white board cells, while movement rules apply only to black cells. A correct GDL simulator therefore drops white-cell facts after the initial state. In contrast, LLMs often retain these facts in subsequent states, despite the absence of rules that could reintroduce them, leading to redundant state descriptions. Similarly, in Tasks 3 and 4, where models must begin with inferring the initial state, they sometimes include constant facts, which are not part of the state representation.

\paragraph{Violation of strict movement constraints.}
In Tasks 2 and 4 involving action generation, models occasionally violate the constraints encoded in the rules. For example, in \texttt{battlebrushes}, where movement is restricted to four cardinal directions, models tend to propose diagonal moves, even though no rule permits them. This might be caused by the compositional complexity -- the rule for legality uses the condition named \texttt{adjacent} and the rules for \texttt{adjacent} are defined separately. 

\paragraph{Formatting errors.}
Despite explicit instructions and relaxed tokenization rules (see supplementary material), occasional formatting mistakes still occur, even for the strongest models. For example, in \texttt{snakeAssembly}, outputs such as \texttt{score(blue,5)} appear instead of the \texttt{(score blue 5)}. A similar type of error was observed for \texttt{battlebrushes} in which the strongest model (Gemini 2.5 Pro) repeatedly provided syntactically incorrect outputs, despite proper recognition of all relevant facts. While we tolerated the use of different separators, incorrect use of brackets violates the GDL syntax too heavily and results in formally invalid states.

\subsection{Obfuscation}
\label{sec:obfuscation}

We evaluate the impact of symbol obfuscation on the Multistep state generation (Task 3, $n=5$).
Table~\ref{tab:obfuscation} summarizes the results. Overall, performance decreases under obfuscation, but the degradation is mild. This indicates that \textit{the models can still perform multistep symbolic reasoning even when surface-level tokens no longer carry their original meaning}. In particular, the fact that performance remains relatively high despite obfuscation suggests that the probable presence of similar game descriptions in the models' pretraining data was not a prerequisite for their successes.

Detailed analysis reveals that the specific type of obfuscation matters. The 'dictionary words' variant proved to be the hardest, suggesting that models get confused when words carry real-world meanings that conflict with game rules. In contrast, using 'random strings' resulted in the highest performance among the obfuscated settings, even outperforming generic `placeholder terms'. This indicates that stripping away all linguistic clues prevents \textit{semantic leakage}, forcing the model to rely entirely on the provided logic. Generic placeholders (term1, term2) effectively remove semantic bias, but they introduce high lexical similarity, which may affect the model's ability to distinguish between numerous state variables. In contrast, random strings act as distinct, unique identifiers, which likely helps the model in tracking symbols across multiple steps by minimizing the risk of token confusion.

While preserving the original semantic meaning still provides the greatest advantage (cf. last row of Table~\ref{tab:obfuscation}), the results confirm that strong models can reason symbolically without needing familiar words to guide them.
Statistical significance of these comparisons was verified using Wilcoxon signed-rank tests. Detailed results are provided in the supplementary material.

\begin{table}[ht]
\centering
\caption{Results for the obfuscation method using placeholder terms (term1, term2, \dots) for selected games. The average results over all 35 games are included for all three obfuscation methods: placeholder terms, dictionary words, random strings, and for a baseline no-obfuscated game definition.
}
\label{tab:obfuscation}
\setlength{\tabcolsep}{3pt}
\begin{tabular}{lcccccccc}
\toprule
 & \multicolumn{2}{c}{Gemini 2.5 Pro} & \multicolumn{2}{c}{Gemini 2.5 Flash} & \multicolumn{2}{c}{GPT-OSS 120B} & \multicolumn{2}{c}{Llama 3.3 70B} \\
\cmidrule(lr){2-3} \cmidrule(lr){4-5} \cmidrule(lr){6-7} \cmidrule(lr){8-9}
Game & JI & \%S & JI & \%S & JI & \%S & JI & \%S \\
\midrule
battlebrushes & 0.169 & 0.150 & 0.465 & 0.250 & 0.127 & 0.000 & 0.069 & 0.000 \\
bomberman2p & 0.970 & 0.650 & 0.889 & 0.250 & 0.446 & 0.050 & 0.647 & 0.000 \\
checkers & 0.996 & 0.900 & 0.949 & 0.350 & 0.048 & 0.000 & 0.447 & 0.000 \\
chess & 0.941 & 0.550 & 0.904 & 0.150 & 0.192 & 0.000 & 0.347 & 0.000 \\
mummymaze2p & 0.995 & 0.600 & 0.972 & 0.500 & 0.607 & 0.000 & 0.746 & 0.000 \\
pacman3p & 0.947 & 0.750 & 0.956 & 0.300 & 0.231 & 0.000 & 0.872 & 0.000 \\
platformJumpers & 0.525 & 0.150 & 0.050 & 0.050 & 0.450 & 0.000 & 0.000 & 0.000 \\
snake\_2009\_big & 0.864 & 0.200 & 0.892 & 0.100 & 0.110 & 0.000 & 0.019 & 0.000 \\
\midrule
\textbf{Avg (term placeholder)} & \textbf{0.822} & \textbf{0.583} & 0.783 & 0.427 & 0.298 & 0.083 & 0.292 & 0.010 \\
\textbf{Avg (dictionary words)}  & \textbf{0.852} & \textbf{0.523} & 0.774 & 0.313 & 0.237 & 0.014 & 0.262 & 0.001 \\
\textbf{Avg (random strings)} & \textbf{0.898} & \textbf{0.643} & 0.822 & 0.411 & 0.345 & 0.079 & 0.336 & 0.059 \\
\bottomrule
\textbf{Avg (no obfuscation)} & \textbf{0.865} & \textbf{0.734} & 0.843 & 0.591 & 0.688 & 0.389 & 0.493 & 0.133 \\
\bottomrule
\end{tabular}
\end{table}

\subsection{Correlation with Game Properties}

Table~\ref{tab:correlations} 
presents 
a linear dependence between game features and model performance using the Pearson correlation coefficient ($r$). Performance is measured using the $\%S$ metric averaged across all tasks, separately for each model. The correlation analysis reveals that the notion of "difficulty" in logical reasoning is not universal but highly dependent on the model's capacity and architecture. We observed distinct reasoning bottlenecks for each evaluated model. 

\begin{table}[ht]
\centering
\caption{Top-10 correlations averaged across models between game properties and model performance. Rule's `H-Index' is defined as max $H$ s.t. $H$ rules have $\ge H$ conditions. 
`Converging property` indicates that state facts are never deleted, only added. Full results are presented in supplementary material. The highest absolute value for each model is \textbf{bolded}.}
\label{tab:correlations}
\setlength{\tabcolsep}{3pt}
\begin{tabular}{lcccc}
\toprule
Property & Gemini 2.5 Pro & Gemini 2.5 Flash & GPT-OSS 120B & Llama 3.3 70B \\
\midrule
Converging property & 0.45 & 0.53  & 0.46 & \textbf{0.71} \\
Total rules\_NEXT & -0.49 & \textbf{-0.60} & -0.43 & -0.45 \\
Total constant facts\_NEXT & -0.47 & -0.53 & \textbf{-0.49} & -0.44 \\
Rules 'H-Index'\_NEXT & -0.48 & -0.49 & -0.42 & -0.50 \\
Total conditions\_NEXT & -0.49 & -0.54 & -0.41 & -0.44 \\
Max conditions per rule\_NEXT & -0.48 & -0.49 & -0.42 & -0.39 \\
Max rule depth\_NEXT & \textbf{-0.53} & -0.51 & -0.32 & -0.32 \\
Avg rule depth\_NEXT & -0.52 & -0.46 & -0.26 & -0.38 \\
Top level rules\_NEXT & -0.36 & -0.48 & -0.35 & -0.39 \\
Top level rules\_LEGAL & -0.40 & -0.48 & -0.38 & -0.29 \\
\bottomrule
\end{tabular}%
\end{table}

For the strongest model, Gemini 2.5 Pro, performance is primarily constrained by the nested complexity of the problem. The strongest negative correlation are \textit{Max rule depth NEXT} ($r=-0.53$) and \textit{Avg rule depth NEXT} ($r=-0.52$). This suggests that the model is particularly sensitive to recursive logic and struggles to maintain consistency across deep, compositional inference chains. 

In contrast, the more lightweight Gemini 2.5 Flash appears to be limited by the length of game definitions rather than their structural depth. Its performance drops most significantly with an increase in \textit{Total rules NEXT} ($r=-0.60$), indicating that a large search space overwhelms the model's reasoning mechanism, even if the individual logical steps are shallow. A similar situation is observed for GPT-OSS 120B.

Llama 3.3 70B shows strong positive correlation with the \textit{Converging} property ($r=0.71$) which may indicate that model's reasoning capabilities are limited by the fundamental tractability of the game dynamics itself not by specific logical features (depth or volume) as is the case of other models.

\section{Conclusions}
\label{sec:conclusions}

We demonstrated that the GGP/GDL framework can serve as a testbed for evaluating logical reasoning in large language models thanks to the combination of formally specified dynamics, a diverse collection of games, and an exact GDL interpreter serving as ground truth.

We evaluated four contemporary LLMs and established an experimental setup that can be directly extended to additional models in the future. Our results show that the strongest models have made substantial progress in forward symbolic reasoning and generally can accurately generate successor states even in games with relatively complex state representations.

Obfuscation experiments further indicate that models are able to perform multistep logical reasoning even when identifiers carry no linguistic meaning and when the likelihood of overlap with training data is minimal. This suggests that they rely on structural patterns in the rules rather than on surface-level semantics alone.

At the same time, several fundamental limitations remain. Performance drops for longer reasoning chains in which both actions and states are generated, where even small local errors quickly compound and lead to incorrect final states. In addition, games with deeper compositional structure and a larger number of interacting rules consistently yield lower accuracy.

Our analysis of common mistakes reveals that we should not treat models as faithful symbolic simulators in critical applications. Instead, their outputs should be verified, or the interaction should be structured iteratively, for example by explicitly checking understanding of the rules, validating intermediate states, or querying for potential inconsistencies.

\paragraph{Limitations.} Since, to our knowledge, this is the first approach to assessing LLMs in the GDL setting, some research paths remained unexplored. Future work should extend this evaluation to a broader range of models and investigate performance under systematically controlled game parameters. For instance, designing parameterizable game families would enable ablation-style studies of how specific complexity factors affect reasoning performance, providing more detailed insight into the limits and scaling behavior of symbolic 

\section*{Source code} 
The source code and all related files are publicly available at
\url{https://doi.org/10.5281/zenodo.18717431}. Further details are provided in the supplementary material.

\bibliographystyle{unsrt}  
\bibliography{references}  

\clearpage
\onecolumn

\begin{center}
  \textbf{\LARGE Reasoning Capabilities of Large Language Models. \\
Lessons Learned from General Game Playing\\
--- Supplementary Material ---}
  \vspace{0.5cm}
\end{center}

\setcounter{equation}{0}
\setcounter{figure}{0}
\setcounter{table}{0}
\setcounter{page}{1}
\setcounter{section}{0}

\renewcommand{\theequation}{S\arabic{equation}}
\renewcommand{\thefigure}{S\arabic{figure}}
\renewcommand{\thetable}{S\arabic{table}}
\renewcommand{\thesection}{S\arabic{section}}

All files associated with this paper are publicly available at:

\begin{center}
\url{https://doi.org/10.5281/zenodo.18717431}
\end{center}
\section{File Structure of the Supplementary Material}

\begin{itemize}
  \item \texttt{Code}
  \begin{itemize}
    \item \texttt{GDLProject}
    \item \texttt{ObfuscatorGDL}
    \item \texttt{Parser}
    \item \texttt{llm\_runner.py}
  \end{itemize}
  \item \texttt{No-Obfuscation}
  \begin{itemize}
    \item \texttt{Games}
    \item \texttt{Problem Instances}
    \item \texttt{Results}
  \end{itemize}
  \item \texttt{Obfuscation-1}
  \begin{itemize}
    \item \texttt{Games}
    \item \texttt{Problem Instances}
    \item \texttt{Results}
    \item \texttt{encoding.txt}
  \end{itemize}
  \item \texttt{Obfuscation-2}
  \begin{itemize}
    \item \texttt{Games}
    \item \texttt{Problem Instances}
    \item \texttt{Results}
    \item \texttt{encoding.txt}
  \end{itemize}
  \item \texttt{Obfuscation-3}
  \begin{itemize}
    \item \texttt{Games}
    \item \texttt{Problem Instances}
    \item \texttt{Results}
    \item \texttt{encoding.txt}
  \end{itemize}
\end{itemize}

\paragraph{Problem Instances.}
The \texttt{Problem Instances} directories are divided into two task-specific subfolders:
\texttt{Next State and Legal Actions Generation} and \texttt{Multistep State Generation}.
The former contains benchmark instances shared by both the Next state generation and Legal action generation tasks, since both tasks use the same input in the form of the current game state.
Each benchmark instance includes the fields \texttt{game\_state} (used by both tasks), \texttt{move} (used by the next-state generation task), and ground-truth outputs for both tasks: \texttt{legal\_moves} and \texttt{next\_state}.

The \texttt{Multistep State Generation} folder contains benchmark instances for Multistep state generation.
Each instance is provided as an action sequence of length up to $n=15$.
For experiments with a given horizon $n$, only the first $n$ actions from the sequence are used.
Ground-truth states are not stored explicitly; instead, they are computed dynamically from the action sequences using the \texttt{GDLProject} program.
The Multistep action--state generation task does not use predefined benchmark instances, as the corresponding action sequences are generated directly by the LLMs.

\paragraph{Results.}
The \texttt{Results} directories contain a two-level folder hierarchy.
The first level corresponds to the task (e.g., \texttt{Multistep State Generation n=5}), while the second level corresponds to the evaluated model (e.g., \texttt{gemini-2.5-pro}) and contains the outputs produced by that model.

\paragraph{Code.}
The \texttt{GDLProject} directory contains the core program used for all tasks requiring GDL interpretation, including benchmark generation and verification of LLM outputs.
The project is implemented in C\# using Microsoft Visual Studio.
The main entry point is located in \texttt{MainNode/MainNodeProgram.cs}.
Before running the program, users should review the lines marked with \texttt{TODO} in this file.
The project was developed for internal use and is not yet fully documented; upon acceptance of the paper, we plan to provide comprehensive user documentation and a refined API.

The \texttt{ObfuscatorGDL} directory contains a utility for obfuscating and deobfuscating GDL files using a specified encoding.
The \texttt{Parser} directory provides a standalone parsing library used internally by \texttt{GDLProject}.
Finally, \texttt{llm\_runner.py} is a Python script that implements the interface used to interact with the evaluated large language models.

\section{Prompting Methodology}
To rigorously evaluate the logical reasoning capabilities of LLMs within the GGP framework, we developed an automated evaluation pipeline. Our interaction strategy relies on a zero-shot prompting approach, where models are required to interpret raw GDL code without intermediate translation into natural language. This design choice effectively tests the model's ability to act as a symbolic interpreter rather than relying on linguistic pattern matching.

We constructed four distinct prompt templates corresponding to the experimental tasks described in Section 4.1 of the main text. Each prompt adheres to a consistent structure consisting of the following components:

\begin{itemize}
    \item \textbf{System Persona.} We explicitly instruct the model to adopt the role of a "game logic expert," setting the context for formal symbolic manipulation.
    \item \textbf{Game Definition.} The complete, unmodified GDL description of the game is provided in a delimited block. This ensures the model has access to all rules, facts, and keywords required for inference.
    \item \textbf{State and Action Context.} Depending on the task, we provide the current state of the game (as a set of facts) and optionally list of the moves to be executed. For multistep experiment all moves were passed in one prompt.
    \item \textbf{Output Constraints.} To enable automated parsing and comparison against the ground truth simulator, we enforce a strict JSON output format. Models are instructed to output only the requested data (e.g., the set of next facts or legal actions) without auxiliary explanations or reasoning traces.
\end{itemize}

All interactions were conducted with a temperature setting of $0.2$. This low temperature was chosen to minimize stochasticity and hallucination, encouraging the models to adhere strictly to the logical constraints defined in the GDL rules. 

\subsection{Prompt Templates}
\label{sec:prompts}

This section provides the exact prompt templates used for interacting with the Large Language Models across the four experimental tasks. In the templates below, placeholders enclosed in braces (e.g., \texttt{\{game\_definition\}}, \texttt{\{game\_state\}}) are replaced programmatically with the actual data for each test instance.

\subsubsection{Task 1: Next state generation}

\begin{quote}
\small\ttfamily
You are a game logic expert. Your task is to predict the next game state.

Here is the game definition in GDL (Game Description Language). GDL was used in General Game Playing competition. Ignore the Init part; the current state will be provided later:
\\
- - - GAME DEFINITION - - -
\\
\{game\_definition\}
\\
- - - - - - - - - - - - - - - - - - - - - - -

Here is the current game state:
\\
- - - GAME STATE - - -
\\
\{game\_state\}
\\
- - - - - - - - - - - - - - - - - -

The following move is being executed:
\\
- - - MOVE - - -
\\
\{move\}
\\
- - - - - - - - - - - -

What will be the **exact** game state after this move?
Respond **only** in JSON format using the following fields:
\\
- "llm\_state": string, the complete new game state in the same format as the input state, each fact separated by new line symbol

Do not add any explanations, comments, or markdown formatting.
Your response must start with \{.
\end{quote}

\subsubsection{Task 2: Legal actions generation}

\begin{quote}
\small\ttfamily
You are a game logic expert. Your task is to determine all legal moves available for all players in the current state.

Here is the game definition in GDL (Game Description Language):
\\
- - - GAME DEFINITION - - -
\\
\{game\_definition\}
\\
- - - - - - - - - - - - - - - - - - - - - - -

Here is the current game state:
\\
- - - GAME STATE - - -
\\
\{game\_state\}
\\
- - - - - - - - - - - - - - - - - -

List **all** legal moves for this state based on the GDL rules.
Respond **only** in JSON format using the following field:
\\
- "llm\_legal\_moves": string which contains all possible valid GDL moves (role + action) separated by new line symbol, each move should be in round brackets

Do not add any explanations, comments, or markdown formatting. Don't put GDL syntactic elements like "legal", "does" in the moves.
Your response must start with \{.
\end{quote}

\subsubsection{Task 3: Multistep state generation}
\begin{quote}
\small\ttfamily
You are a game logic expert. Your task is to predict the game state after a specific sequence of moves.

Here is the game definition in GDL (Game Description Language). 
The game starts from the initial state defined in this GDL (look for 'init' relations).
\\
- - - GAME DEFINITION - - -
\\
\{game\_definition\}
\\
- - - - - - - - - - - - - - - - - - - - - - -

Starting from the initial state defined above, apply the following sequence of \{n\} moves in order:
\\
- - - MOVE SEQUENCE - - -
\\
\{move\_sequence\}
\\
- - - - - - - - - - - - - - - - - - - - -

What will be the **exact** game state after these \{n\} moves have been executed?
Respond **only** in JSON format using the following fields:
\\
- "llm\_state": string, the complete new game state in GDL format, each fact separated by new line symbol.

Do not add any explanations, comments, or markdown formatting.
Your response must start with \{.
\end{quote}

\subsubsection{Task 4: Multistep action-state generation}

\begin{quote}
\small\ttfamily
You are a game logic expert. Your task is to play the game for \{n\} steps and predict the resulting state.
Moves don't need to be optimal, they can be random but each of them have to be legal.

Here is the game definition in GDL (Game Description Language).
The game starts from the initial state defined in this GDL (look for 'init' relations).
\\
- - - GAME DEFINITION - - -
\\
\{game\_definition\}
\\
- - - - - - - - - - - - - - - - - - - - - - -

**Task:**
\\
1. Starting from the initial state, generate a valid sequence of \{n\} moves.
\\
2. Calculate the exact game state after these \{n\} moves.

**Move Format:**
Your generated moves must follow this specific string format (example from this game):
\\
\{move\_example\}

**Output:**
Respond **only** in JSON format using the following structure:
\\
\{\{
    "moves": [
        \{\{ "step": "0", "joint\_move": "..." \}\},
        \{\{ "step": "1", "joint\_move": "..." \}\}
    ],
    "llm\_state": "string containing the complete new game state in GDL format"
\}\}

Do not add any explanations, comments, or markdown formatting.
Your response must start with \{.
\end{quote}

\section{GDL Definitions of Games}

GDL definitions were downloaded from the Stanford~\cite{stanford-gdl} and Dresden~\cite{dresden-gdl} game repositories. 
All 86 games are available in the ZIP archive attached as Supplementary Material under the directories:

\begin{enumerate}
    \item \texttt{No-Obfuscation\textbackslash Games}: original, unmodified GDL definitions.
    \item \texttt{Obfuscation-1\textbackslash Games}: definitions using the \emph{placeholder-term} obfuscation variant.
    \item \texttt{Obfuscation-2\textbackslash Games}: definitions using the \emph{dictionary-words} obfuscation variant.
    \item \texttt{Obfuscation-3\textbackslash Games}: definitions using the \emph{random-strings} obfuscation variant.
\end{enumerate}

The directory structure is organized such that the top-level folder (e.g., \texttt{Obfuscation-1}) aggregates all information subject to a given obfuscation scheme. Subfolders at the next level group different types of content. For example, as mentioned before, \texttt{Games} contains the corresponding GDL game definitions.

\subsection{Obfuscation}

In each obfuscated directory, a file named \texttt{encoding.txt} is provided. It contains lines of the form:
\begin{quote}
\small\ttfamily
<original\_symbol> <obfuscated\_symbol>
\end{quote}

Each line specifies a one-to-one mapping from an original symbol to its obfuscated counterpart, separated by a single space. Symbols do not contain whitespace. The encoding is applied globally within a directory, i.e., the same original symbol is consistently mapped to the same obfuscated symbol across all games.

Below, we summarize the obfuscation methods used and provide illustrative examples:
\begin{enumerate}
    \item \textbf{Placeholder terms}: Each non-reserved symbol is replaced by \texttt{term<i>}, where \texttt{i} is a consecutive index incremented for every unique non-reserved symbol encountered. Example encoding lines:
    \begin{verbatim}
cell term12
?color ?term13
opponent term14
    \end{verbatim}

    \item \textbf{Dictionary words}: Symbols are mapped to words drawn from a family-friendly dictionary focusing on nature (e.g., weather, terrain, plants, non-violent animals), geography (neutral landforms and directions), objects (tools, furniture, materials), and abstract but non-sensitive concepts (e.g., shape, size, motion, time). Example encoding lines:
    \begin{verbatim}
cell openvessel
?color ?horizoncrystal
opponent calmcompass
    \end{verbatim}
    The corresponding \texttt{encoding.txt} file can be inspected to obtain an overview of the generated vocabulary.

    \item \textbf{Random strings}: Each unique symbol is mapped to a unique random alphanumeric string of length 5--8, consisting of upper- and lower-case letters and digits. Strings were generated using a brute-force procedure with retries to avoid collisions. Example encoding lines:
    \begin{verbatim}
cell AoinIj6
?color ?ZVsoQ
opponent Uivv5E
    \end{verbatim}
\end{enumerate}

Reserved symbols include GDL keywords as well as property names used in experimental JSON logs (e.g., task or model identifiers). These symbols are left unobfuscated so that de-obfuscation can be safely applied by simply transforming all words in a file using the provided \texttt{encoding.txt}. Property names used in JSON logs do not appear in the game descriptions.

\subsection{Game Selection Procedure }

Let $S^i$ denote the $i$-th best average score obtained in preliminary experiments on the \emph{nextstate} prediction task, computed over 20 repetitions per game. In particular, $S^1$ denotes the score achieved by the best-performing model for a given game.

From the full set of 86 game descriptions, we selected candidate games by choosing the top 10 instances according to each of the following criteria, followed by duplicate removal:

\begin{itemize}
\item \textbf{AvgFactCount}: Games with the largest average number of facts in their state descriptions.
\item \textbf{StateDiff}: Games with the largest average number of facts added or removed between consecutive states.
\item \textbf{HighEnd}: Games maximizing
\[
\sum_i S^i + \prod_i S^i.
\]
\item \textbf{LowEnd}: Games minimizing
\[
\sum_i S^i + \prod_i S^i.
\]
\item \textbf{Variance}: Games maximizing
\[
(S^1 - S^2) + (S^1 - S^3).
\]
\end{itemize}

After removing duplicates, i.e., games that appeared in multiple top-10 lists and two games with identical descriptions (\texttt{pawnWhopping} and \texttt{pawnWhoppingCorrected}), a total of 35 unique games remained. These games are listed in Table~\ref{tab:games}. The remaining games, i.e., those excluded after the preliminary experiments, are listed in Table~\ref{tab:excluded-games}.

\begin{table*}[ht]
\centering
\small
\caption{Chosen games for the empirical experiments.}
\label{tab:games}
\begin{tabular}{cccccc}
\toprule
 & \multicolumn{5}{c}{Place in TOP 10} \\
\cmidrule{2-6}
Game & AvgFactCount & StateDiff & HighEnd & LowEnd & Variance \\
\midrule
1reversi2                   &  &  &  &  &  5 \\
battlebrushes               &  8 & 6 &  &  &  \\
beatMania                   &  &  4 &  &  &  \\
bomberman2p                 &  7 &  &  &  & 4 \\
bomberman2p\_InvertedRoles  &  6 & 10 & & 8  & 3  \\
buttons                     &  &  & 4 &  &  \\
checkers                    &  9 &  &  &  &  \\
checkers-mustjump           &  10 &  &  & 6 &  \\
checkers-newgoals           &  &  &  &  &  9 \\
checkersSmall               &  &  &  &  9 &  6 \\
checkersTiny                &  &  &  &  2 &  \\
chess                       &  &  &  &  5 &  8\\
chineseCheckers3            &  &  &  6 &  &  \\
cittaceot                   &  &  1 &  &  &  \\
connectfour                 &  &  &  9 &  &  \\
connectFourSuicide          &  &  &  10 &  &  \\
dotsAndBoxes                &  &  &  5 &  &  \\
dotsAndBoxesSuicide         &  &  &  3 &  &  \\
farmers                     &  &  5 &  & 10 &  \\
fighter                     &  &  &  &  &  1\\
god                         &  &  3 &  &  7 &  \\
mummymaze2p                 &  1 &  &  &  &  \\
othello-comp2007            &  &  &  &  3 & 7  \\
othellosuicide              &  &  &  &  4 &  2 \\
pacman3p                    &  4 &  &  &  &  \\
pawnWhopping                &  &  & 7  &  &  \\
platformJumpers             &  3 &  &  &  &  \\
qyshinsu                    &  & 7  &  & 1 &  \\
rendezvous\_asteroids       &  & 9 &  &  &  \\
rubikscube                  &  & 2 &  &  & 10 \\
snake\_2009\_big            &  5 &  &  &  &  \\
snakeAssemblit              &  &  8 &   &  &  \\
ticTacToeLarge              &  &  & 2 &  &  \\
ticTacToeLargeSuicide       &  &  & 1 &  &  \\
wallmaze                    &  2 &  &  &  &  \\
\bottomrule
\end{tabular}
\end{table*}

In the main text, we report per-game scores for the eight games with the highest average fact counts in the state representation, excluding \texttt{wallmaze} (2nd), \texttt{bomberman2p\_InvertedRoles} (6th), and \texttt{checkers-mustjump} (10th) because they are very similar to \texttt{mummymaze2} (1st), \texttt{bomberman2p} (7th), and \texttt{checkers} (9th), respectively. This leads to the inclusion of \texttt{chess}, which ranks 11th according to this metric.

\begin{table*}[ht]
\centering
\small
\caption{Games from the full set of 86 instances that \textbf{were not selected} for the final experimental setup of 35 instances.}
\label{tab:excluded-games}
\begin{tabular}{lll}
\toprule
2pffa\_zerosum        & breakthroughSmallHoles & nineBoardTicTacToe \\
2pttc                & breakthroughSuicide    & pawnWhoppingCorrected \\
3conn3               & breakthroughWalls      & pawnToQueen \\
3pttc                & cephalopodMicro         & peg \\
4pttc                & checkersBarrelNoKings  & pegEuro \\
bidding-tictactoe    & chinookDisjunctive      & pentago \\
biddingTicTacToe\_10coins & circlesolitaire   & pentago\_2008 \\
blocker              & connectFourLarge        & pentagoSuicide \\
breakthrough         & dualConnect4            & pilgrimage \\
breakthroughHoles    & eightPuzzle             & quarto \\
breakthroughSmall    & escortLatch             & roshambo2 \\
farmingQuandries     & four\_way\_battle       & sheep\_and\_wolf \\
golden\_rectangle    & hanoi                   & skirmish \\
knightstour          & knightThrough           & stratego \\
knightwar            & lightsout               & tictactoe \\
logistics            & max\_knights            & tictactoe\_3player \\
ttcc4\_2player       & tictactoeNotOr          & withConviction \\
\bottomrule
\end{tabular}
\end{table*}

\section{Games Properties}

This section summarizes the game properties used in our analysis. The values of those properties which have the highest correlation with models' performance, computed for each game, are reported in Table~\ref{tab:all_games_features}. Below, we provide definitions and explanations of each property to clarify their meaning and how they are derived from the GDL game descriptions.

Most complexity-related measures are computed over subsets of rules responsible for deriving a given concept. These subsets are identified by a suffix:
\begin{itemize}
    \item \texttt{\_NEXT}: rules defining the next game state
    \item \texttt{\_LEGAL}: rules defining available actions
\end{itemize}

Below we describe features defined over the \texttt{\_NEXT} or \texttt{\_LEGAL} rule subsets:

\begin{itemize}
    \item \textbf{Avg / Max arguments per condition} -- the average or maximum (over rules) number of arguments (variables and constants) appearing in rule conditions.
    \item \textbf{Avg / Max conditions per rule} -- the average or maximum number of conditions per rule.
    \item \textbf{Avg / Max rule depth} -- the average or maximum depth of rules, measured by analyzing dependencies top-down; encountering a condition that requires firing another rule increases the depth by one.
    \item \textbf{Conditions 'H-Index'} -- the largest value $H$ such that there exists a condition with at least $H$ arguments.
    \item \textbf{Negative conditions} -- the number of negated conditions.
    \item \textbf{Or conditions} -- the number of conditions composed of multiple subconditions connected by the \emph{or} operator. Standard conditions are not compound and are implicitly combined within a rule using \emph{and}.
    \item \textbf{Recurrent rules} -- the number of recursive rules.
    \item \textbf{Rules 'H-Index'} -- the largest value $H$ such that there exist $H$ rules with at least $H$ conditions each.
    \item \textbf{Top level rule types} -- the number of object types (GDL term names) for which top-level rules are defined. For \texttt{LEGAL}, this corresponds to the number of distinct action types (e.g., \texttt{move}, \texttt{capture}), while for \texttt{NEXT} it corresponds to the number of state descriptor types (e.g., \texttt{cell}, \texttt{turn\_number}).
    \item \textbf{Top level rules} -- the total number of defined top-level rules (i.e., \texttt{LEGAL} or \texttt{NEXT}). This number is at least as large as the number of top-level rule types, since a single action or state type may be defined by multiple rules.
    \item \textbf{Total conditions} -- the total number of conditions across all rules.
    \item \textbf{Total constant facts} -- the total number of constant facts in the GDL description that are used by the rules.
    \item \textbf{Total rules} -- the total number of rules (as always: restricted to those involved in computing \texttt{LEGAL} or \texttt{NEXT}, respectively).
\end{itemize}

The following features are global, game-level properties:

\begin{itemize}
    \item \textbf{"Connection game" elements?} -- whether the game contains mechanics based on connecting board positions, as in Tic-Tac-Toe.
    \item \textbf{2D Board?} -- whether the game includes a two-dimensional board.
    \item \textbf{Avg FactCount} -- the empirical average number of facts representing a game state.
    \item \textbf{Converging?} -- a binary feature (0/1). If the game is converging, once a fact appears in the game state, it persists until the end of the game (e.g., objects cannot be moved or removed).
    \item \textbf{Diff Heuristic} -- the average number of facts that are added or removed when comparing states before and after applying a \texttt{next} rule.
    \item \textbf{Has truly simultaneous moves?} -- although GDL formally specifies simultaneous moves, not all moves are semantically meaningful. For example, the \texttt{no\_op} action (no operation) is used to model turn-based games and has no effect on the state. This feature indicates whether simultaneous moves are genuinely meaningful and jointly affect the game state, rather than serving as artificial placeholders.
    \item \textbf{No. of Players} -- the number of players.
    \item \textbf{Offboard resource management?} -- whether the game involves managing abstract resources (e.g., money, grain, energy) that are not represented as board pieces.
    \item \textbf{Point counting?} -- whether winning is determined by accumulating points rather than, for example, eliminating all opponent pieces.
\end{itemize}

\section{Detailed results}

This section provides comprehensive tabular results for all experimental tasks and all evaluated games. The tables extend the summarized results reported in the main text by presenting per-game performance, additional reasoning horizons, and detailed obfuscation results for all variants in Multistep state generation task with $n=5$.

\subsection{Next state generation}

\noindent Table~\ref{tab:next_state2} reports full Next state generation results for all 35 games, extending the next-state generation results shown for selected games in Table~1 in the main text.

\begin{table*}[ht]
\centering
\small
\caption{Next state generation results across all games.}
\label{tab:next_state2}
\begin{tabular}{lcccccccc}
\toprule
 & \multicolumn{2}{c}{Gemini 2.5 Pro} & \multicolumn{2}{c}{Gemini 2.5 Flash} & \multicolumn{2}{c}{GPT-OSS 120B} & \multicolumn{2}{c}{Llama 3.3 70B} \\
\cmidrule(lr){2-3} \cmidrule(lr){4-5} \cmidrule(lr){6-7} \cmidrule(lr){8-9}
Game & JI & \%S & JI & \%S & JI & \%S & JI & \%S \\
\midrule
1reversi2 & 1.000 & 1.000 & 0.998 & 0.950 & 0.996 & 0.950 & 0.823 & 0.000 \\
battlebrushes & 0.997 & 0.950 & 0.997 & 0.950 & 0.988 & 0.750 & 0.945 & 0.100 \\
beatMania & 1.000 & 1.000 & 1.000 & 1.000 & 0.998 & 0.950 & 0.242 & 0.000 \\
bomberman2p & 0.995 & 0.950 & 0.972 & 0.700 & 0.932 & 0.500 & 0.946 & 0.350 \\
bomberman2p\_InvertedRoles & 0.976 & 0.800 & 0.963 & 0.650 & 0.920 & 0.500 & 0.909 & 0.050 \\
buttons & 1.000 & 1.000 & 1.000 & 1.000 & 1.000 & 1.000 & 0.682 & 0.350 \\
checkers & 1.000 & 1.000 & 0.999 & 0.950 & 0.994 & 0.850 & 0.956 & 0.250 \\
checkers-mustjump & 1.000 & 1.000 & 0.999 & 0.950 & 0.991 & 0.800 & 0.939 & 0.200 \\
checkers-newgoals & 1.000 & 1.000 & 1.000 & 1.000 & 0.994 & 0.900 & 0.931 & 0.150 \\
checkersSmall & 1.000 & 1.000 & 1.000 & 1.000 & 0.970 & 0.900 & 0.913 & 0.400 \\
checkersTiny & 1.000 & 1.000 & 0.995 & 0.950 & 0.991 & 0.950 & 0.894 & 0.450 \\
chess & 0.998 & 0.850 & 0.998 & 0.850 & 0.996 & 0.800 & 0.973 & 0.150 \\
chineseCheckers3 & 1.000 & 1.000 & 1.000 & 1.000 & 1.000 & 1.000 & 0.984 & 0.650 \\
cittaceot & 0.951 & 0.950 & 1.000 & 1.000 & 1.000 & 1.000 & 0.296 & 0.000 \\
connectfour & 1.000 & 1.000 & 1.000 & 1.000 & 1.000 & 1.000 & 0.979 & 0.850 \\
connectFourSuicide & 1.000 & 1.000 & 1.000 & 1.000 & 1.000 & 1.000 & 0.982 & 0.850 \\
dotsAndBoxes & 0.974 & 0.850 & 0.963 & 0.750 & 0.974 & 0.850 & 1.000 & 1.000 \\
dotsAndBoxesSuicide & 0.994 & 0.950 & 0.990 & 0.900 & 0.994 & 0.950 & 0.981 & 0.900 \\
farmers & 0.992 & 0.950 & 0.910 & 0.800 & 0.962 & 0.800 & 0.492 & 0.000 \\
fighter & 1.000 & 1.000 & 0.926 & 0.600 & 0.838 & 0.100 & 0.677 & 0.050 \\
god & 0.979 & 0.850 & 0.983 & 0.900 & 0.973 & 0.900 & 0.432 & 0.000 \\
mummymaze2p & 1.000 & 1.000 & 0.999 & 0.900 & 1.000 & 1.000 & 0.989 & 0.300 \\
othello-comp2007 & 1.000 & 1.000 & 0.889 & 0.500 & 0.959 & 0.750 & 0.790 & 0.000 \\
othellosuicide & 0.986 & 0.950 & 0.957 & 0.800 & 0.902 & 0.650 & 0.783 & 0.000 \\
pacman3p & 1.000 & 1.000 & 0.994 & 0.750 & 1.000 & 1.000 & 0.989 & 0.650 \\
pawnWhopping & 1.000 & 1.000 & 1.000 & 1.000 & 1.000 & 1.000 & 0.960 & 0.650 \\
platformJumpers & 0.998 & 0.850 & 0.997 & 0.750 & 0.993 & 0.750 & 0.940 & 0.000 \\
qyshinsu & 0.994 & 0.850 & 0.949 & 0.100 & 0.919 & 0.100 & 0.752 & 0.000 \\
rendezvous\_asteroids & 1.000 & 1.000 & 0.866 & 0.550 & 0.976 & 0.900 & 0.570 & 0.050 \\
rubikscube & 1.000 & 1.000 & 0.922 & 0.750 & 0.980 & 0.850 & 0.451 & 0.150 \\
snake\_2009\_big & 0.998 & 0.850 & 0.992 & 0.800 & 0.965 & 0.400 & 0.942 & 0.000 \\
snakeAssemblit & 0.976 & 0.850 & 0.926 & 0.500 & 0.914 & 0.350 & 0.671 & 0.000 \\
ticTacToeLarge & 1.000 & 1.000 & 1.000 & 1.000 & 1.000 & 1.000 & 1.000 & 1.000 \\
ticTacToeLargeSuicide & 1.000 & 1.000 & 1.000 & 1.000 & 1.000 & 1.000 & 1.000 & 1.000 \\
wallmaze & 1.000 & 1.000 & 0.999 & 0.950 & 1.000 & 1.000 & 0.989 & 0.400 \\
\midrule
\textbf{Average} & \textbf{0.995} & \textbf{0.956} & \textbf{0.977} & \textbf{0.836} & \textbf{0.975} & \textbf{0.806} & \textbf{0.823} & \textbf{0.313} \\
\bottomrule
\end{tabular}
\end{table*}

\subsection{Legal actions generation}

\noindent Table~\ref{tab:legal_actions} presents Legal action generation results across all 35 games, corresponding to and extending the results summarized for selected games in Table~2 in the main text.

\begin{table*}[ht]
\centering
\small
\caption{Legal actions generation results across all games.}
\label{tab:legal_actions}
\begin{tabular}{lcccccccc}
\toprule
 & \multicolumn{2}{c}{Gemini 2.5 Pro} & \multicolumn{2}{c}{Gemini 2.5 Flash} & \multicolumn{2}{c}{GPT-OSS 120B} & \multicolumn{2}{c}{Llama 3.3 70B} \\
\cmidrule(lr){2-3} \cmidrule(lr){4-5} \cmidrule(lr){6-7} \cmidrule(lr){8-9}
Game & JI & \%S & JI & \%S & JI & \%S & JI & \%S \\
\midrule
1reversi2 & 0.864 & 0.250 & 0.772 & 0.050 & 0.880 & 0.550 & 0.215 & 0.000 \\
battlebrushes & 1.000 & 1.000 & 1.000 & 1.000 & 1.000 & 1.000 & 0.536 & 0.000 \\
beatMania & 1.000 & 1.000 & 1.000 & 1.000 & 1.000 & 1.000 & 1.000 & 1.000 \\
bomberman2p & 1.000 & 1.000 & 0.928 & 0.500 & 0.965 & 0.850 & 0.525 & 0.000 \\
bomberman2p\_InvertedRoles & 0.988 & 0.900 & 0.919 & 0.550 & 0.983 & 0.850 & 0.533 & 0.000 \\
buttons & 1.000 & 1.000 & 1.000 & 1.000 & 1.000 & 1.000 & 1.000 & 1.000 \\
checkers & 0.954 & 0.800 & 0.813 & 0.300 & 0.201 & 0.050 & 0.004 & 0.000 \\
checkers-mustjump & 1.000 & 1.000 & 0.760 & 0.400 & 0.640 & 0.600 & 0.000 & 0.000 \\
checkers-newgoals & 0.917 & 0.750 & 0.700 & 0.200 & 0.551 & 0.250 & 0.003 & 0.000 \\
checkersSmall & 0.984 & 0.900 & 0.909 & 0.500 & 0.457 & 0.300 & 0.010 & 0.000 \\
checkersTiny & 1.000 & 1.000 & 0.848 & 0.650 & 0.472 & 0.300 & 0.015 & 0.000 \\
chess & 0.783 & 0.150 & 0.311 & 0.000 & 0.371 & 0.000 & 0.143 & 0.000 \\
chineseCheckers3 & 0.866 & 0.600 & 0.687 & 0.350 & 0.881 & 0.650 & 0.372 & 0.000 \\
cittaceot & 1.000 & 1.000 & 1.000 & 1.000 & 0.985 & 0.900 & 0.860 & 0.350 \\
connectfour & 1.000 & 1.000 & 0.994 & 0.950 & 0.972 & 0.750 & 0.420 & 0.000 \\
connectFourSuicide & 1.000 & 1.000 & 0.989 & 0.900 & 0.883 & 0.350 & 0.512 & 0.000 \\
dotsAndBoxes & 1.000 & 1.000 & 0.995 & 0.900 & 0.979 & 0.900 & 0.625 & 0.000 \\
dotsAndBoxesSuicide & 1.000 & 1.000 & 0.993 & 0.800 & 0.999 & 0.950 & 0.633 & 0.000 \\
farmers & 0.995 & 0.950 & 0.995 & 0.950 & 0.995 & 0.950 & 0.425 & 0.000 \\
fighter & 1.000 & 1.000 & 1.000 & 1.000 & 1.000 & 1.000 & 0.598 & 0.000 \\
god & 1.000 & 1.000 & 1.000 & 1.000 & 0.750 & 0.750 & 0.932 & 0.600 \\
mummymaze2p & 1.000 & 1.000 & 1.000 & 1.000 & 0.967 & 0.800 & 0.910 & 0.000 \\
othello-comp2007 & 0.853 & 0.450 & 0.607 & 0.200 & 0.979 & 0.850 & 0.262 & 0.000 \\
othellosuicide & 0.816 & 0.350 & 0.590 & 0.200 & 0.931 & 0.750 & 0.233 & 0.000 \\
pacman3p & 1.000 & 1.000 & 0.954 & 0.750 & 0.982 & 0.900 & 0.483 & 0.000 \\
pawnWhopping & 1.000 & 1.000 & 0.986 & 0.850 & 1.000 & 1.000 & 0.367 & 0.000 \\
platformJumpers & 0.873 & 0.100 & 0.805 & 0.350 & 0.937 & 0.800 & 0.147 & 0.000 \\
qyshinsu & 1.000 & 1.000 & 0.989 & 0.950 & 0.988 & 0.950 & 0.275 & 0.000 \\
rendezvous\_asteroids & 1.000 & 1.000 & 1.000 & 1.000 & 1.000 & 1.000 & 1.000 & 1.000 \\
rubikscube & 1.000 & 1.000 & 1.000 & 1.000 & 1.000 & 1.000 & 1.000 & 1.000 \\
snake\_2009\_big & 1.000 & 1.000 & 0.950 & 0.950 & 0.975 & 0.950 & 0.413 & 0.000 \\
snakeAssemblit & 0.800 & 0.150 & 0.535 & 0.000 & 0.742 & 0.000 & 0.667 & 0.000 \\
ticTacToeLarge & 1.000 & 1.000 & 1.000 & 1.000 & 0.950 & 0.600 & 0.685 & 0.000 \\
ticTacToeLargeSuicide & 1.000 & 1.000 & 0.994 & 0.950 & 0.929 & 0.650 & 0.709 & 0.000 \\
wallmaze & 1.000 & 1.000 & 1.000 & 1.000 & 0.942 & 0.700 & 0.506 & 0.000 \\
\midrule
\textbf{Average} & \textbf{0.963} & \textbf{0.839} & \textbf{0.886} & \textbf{0.691} & \textbf{0.865} & \textbf{0.711} & \textbf{0.486} & \textbf{0.141} \\
\bottomrule
\end{tabular}
\end{table*}


\subsection{Multistep state generation}

Tables~~\ref{tab:multistep_n1} -~\ref{tab:multistep_n10} report Multistep state generation performance with a reasoning horizon of $n=1, 2, 3, 4, 5, 7$, and $10$, respectively, across all games, corresponding to Figure~1 in the main text.


\begin{table*}[ht]
\centering
\small
\caption{Multistep state generation results with $n=1$ steps across all games.}
\label{tab:multistep_n1} 
\begin{tabular}{lcccccccc}
\toprule
 & \multicolumn{2}{c}{Gemini 2.5 Pro} & \multicolumn{2}{c}{Gemini 2.5 Flash} & \multicolumn{2}{c}{GPT-OSS 120B} & \multicolumn{2}{c}{Llama 3.3 70B} \\
\cmidrule(lr){2-3} \cmidrule(lr){4-5} \cmidrule(lr){6-7} \cmidrule(lr){8-9}
Game & JI & \%S & JI & \%S & JI & \%S & JI & \%S \\
\midrule
1reversi2 & 0.800 & 0.800 & 0.883 & 0.800 & 0.796 & 0.750 & 0.714 & 0.000 \\
battlebrushes & 0.900 & 0.900 & 0.902 & 0.750 & 0.161 & 0.000 & 0.656 & 0.000 \\
beatMania & 1.000 & 1.000 & 0.875 & 0.750 & 1.000 & 1.000 & 0.516 & 0.250 \\
bomberman2p & 1.000 & 1.000 & 1.000 & 1.000 & 0.405 & 0.100 & 0.023 & 0.000 \\
bomberman2p\_InvertedRoles & 1.000 & 1.000 & 1.000 & 1.000 & 0.405 & 0.100 & 0.023 & 0.000 \\
buttons & 1.000 & 1.000 & 1.000 & 1.000 & 0.873 & 0.800 & 0.720 & 0.300 \\
checkers & 1.000 & 1.000 & 0.874 & 0.400 & 0.504 & 0.150 & 0.000 & 0.000 \\
checkers-mustjump & 0.899 & 0.850 & 0.596 & 0.450 & 0.994 & 0.750 & 0.000 & 0.000 \\
checkers-newgoals & 0.950 & 0.950 & 0.491 & 0.200 & 0.748 & 0.350 & 0.000 & 0.000 \\
checkersSmall & 0.511 & 0.000 & 0.458 & 0.000 & 0.443 & 0.100 & 0.459 & 0.000 \\
checkersTiny & 0.526 & 0.000 & 0.417 & 0.000 & 0.475 & 0.050 & 0.000 & 0.000 \\
chess & 0.945 & 0.600 & 0.838 & 0.400 & 0.311 & 0.050 & 0.003 & 0.000 \\
chineseCheckers3 & 1.000 & 1.000 & 0.993 & 0.900 & 0.885 & 0.350 & 0.992 & 0.700 \\
cittaceot & 0.919 & 0.850 & 0.950 & 0.950 & 0.950 & 0.700 & 0.857 & 0.000 \\
connectfour & 1.000 & 1.000 & 0.909 & 0.900 & 0.121 & 0.050 & 0.120 & 0.000 \\
connectFourSuicide & 1.000 & 1.000 & 1.000 & 1.000 & 0.341 & 0.300 & 0.063 & 0.000 \\
dotsAndBoxes & 1.000 & 1.000 & 1.000 & 1.000 & 0.917 & 0.900 & 0.840 & 0.200 \\
dotsAndBoxesSuicide & 1.000 & 1.000 & 1.000 & 1.000 & 0.768 & 0.700 & 0.870 & 0.350 \\
farmers & 0.950 & 0.950 & 0.780 & 0.550 & 0.393 & 0.150 & 0.160 & 0.000 \\
fighter & 0.914 & 0.800 & 0.963 & 0.800 & 0.341 & 0.000 & 0.723 & 0.250 \\
god & 0.996 & 0.950 & 0.889 & 0.750 & 0.814 & 0.550 & 0.449 & 0.000 \\
mummymaze2p & 1.000 & 1.000 & 0.900 & 0.900 & 0.787 & 0.050 & 0.992 & 0.500 \\
othello-comp2007 & 1.000 & 1.000 & 0.960 & 0.900 & 0.323 & 0.300 & 0.714 & 0.000 \\
othellosuicide & 1.000 & 1.000 & 0.979 & 0.950 & 0.372 & 0.350 & 0.714 & 0.000 \\
pacman3p & 0.998 & 0.850 & 0.849 & 0.750 & 0.494 & 0.250 & 0.000 & 0.000 \\
pawnWhopping & 1.000 & 1.000 & 0.950 & 0.950 & 0.969 & 0.750 & 0.953 & 0.750 \\
platformJumpers & 0.800 & 0.800 & 0.399 & 0.350 & 0.743 & 0.400 & 0.000 & 0.000 \\
qyshinsu & 0.654 & 0.450 & 0.938 & 0.200 & 0.800 & 0.250 & 0.009 & 0.000 \\
rendezvous\_asteroids & 0.984 & 0.900 & 0.840 & 0.650 & 0.651 & 0.400 & 0.422 & 0.000 \\
rubikscube & 0.948 & 0.850 & 0.736 & 0.650 & 0.453 & 0.250 & 0.531 & 0.250 \\
snakeAssemblit & 0.793 & 0.750 & 0.171 & 0.150 & 0.092 & 0.000 & 0.000 & 0.000 \\
snake\_2009\_big & 1.000 & 1.000 & 0.891 & 0.800 & 0.781 & 0.000 & 0.974 & 0.000 \\
ticTacToeLarge & 1.000 & 1.000 & 1.000 & 1.000 & 0.424 & 0.350 & 0.607 & 0.450 \\
ticTacToeLargeSuicide & 1.000 & 1.000 & 0.422 & 0.200 & 0.483 & 0.350 & 0.286 & 0.000 \\
wallmaze & 0.999 & 0.950 & 0.946 & 0.700 & 0.929 & 0.550 & 0.946 & 0.700 \\
\midrule
\textbf{Average} & \textbf{0.928} & \textbf{0.863} & \textbf{0.823} & \textbf{0.679} & \textbf{0.598} & \textbf{0.347} & \textbf{0.438} & \textbf{0.134} \\
\bottomrule
\end{tabular}
\end{table*}

\begin{table*}[ht]
\centering
\small
\caption{Multistep state generation results with $n=2$ steps across all games.}
\label{tab:multistep_n2} 
\begin{tabular}{lcccccccc}
\toprule
 & \multicolumn{2}{c}{Gemini 2.5 Pro} & \multicolumn{2}{c}{Gemini 2.5 Flash} & \multicolumn{2}{c}{GPT-OSS 120B} & \multicolumn{2}{c}{Llama 3.3 70B} \\
\cmidrule(lr){2-3} \cmidrule(lr){4-5} \cmidrule(lr){6-7} \cmidrule(lr){8-9}
Game & JI & \%S & JI & \%S & JI & \%S & JI & \%S \\
\midrule
1reversi2 & 0.850 & 0.850 & 0.950 & 0.900 & 0.854 & 0.700 & 0.427 & 0.000 \\
battlebrushes & 0.050 & 0.050 & 0.866 & 0.700 & 0.163 & 0.000 & 0.503 & 0.000 \\
beatMania & 1.000 & 1.000 & 1.000 & 1.000 & 1.000 & 1.000 & 0.580 & 0.150 \\
bomberman2p & 1.000 & 1.000 & 0.993 & 0.800 & 0.391 & 0.050 & 0.012 & 0.000 \\
bomberman2p\_InvertedRoles & 1.000 & 1.000 & 0.993 & 0.800 & 0.391 & 0.050 & 0.012 & 0.000 \\
buttons & 1.000 & 1.000 & 1.000 & 1.000 & 0.930 & 0.900 & 0.680 & 0.300 \\
checkers & 1.000 & 1.000 & 0.835 & 0.450 & 0.551 & 0.350 & 0.000 & 0.000 \\
checkers-mustjump & 0.950 & 0.950 & 0.781 & 0.450 & 0.926 & 0.600 & 0.019 & 0.000 \\
checkers-newgoals & 1.000 & 1.000 & 0.897 & 0.800 & 0.632 & 0.400 & 0.000 & 0.000 \\
checkersSmall & 0.535 & 0.050 & 0.393 & 0.000 & 0.484 & 0.000 & 0.334 & 0.000 \\
checkersTiny & 0.528 & 0.000 & 0.500 & 0.000 & 0.524 & 0.250 & 0.383 & 0.000 \\
chess & 0.944 & 0.550 & 0.898 & 0.200 & 0.344 & 0.050 & 0.000 & 0.000 \\
chineseCheckers3 & 0.950 & 0.950 & 0.995 & 0.950 & 0.977 & 0.650 & 0.932 & 0.000 \\
cittaceot & 0.925 & 0.900 & 1.000 & 1.000 & 0.909 & 0.650 & 0.680 & 0.000 \\
connectfour & 1.000 & 1.000 & 1.000 & 1.000 & 0.727 & 0.700 & 0.051 & 0.000 \\
connectFourSuicide & 1.000 & 1.000 & 1.000 & 1.000 & 0.755 & 0.700 & 0.476 & 0.000 \\
dotsAndBoxes & 1.000 & 1.000 & 1.000 & 1.000 & 0.964 & 0.950 & 0.667 & 0.000 \\
dotsAndBoxesSuicide & 1.000 & 1.000 & 1.000 & 1.000 & 0.966 & 0.950 & 0.667 & 0.000 \\
farmers & 0.749 & 0.550 & 0.867 & 0.550 & 0.695 & 0.250 & 0.136 & 0.000 \\
fighter & 0.958 & 0.750 & 0.884 & 0.550 & 0.322 & 0.000 & 0.599 & 0.050 \\
god & 0.964 & 0.850 & 0.864 & 0.450 & 0.736 & 0.400 & 0.574 & 0.000 \\
mummymaze2p & 0.994 & 0.950 & 0.949 & 0.850 & 0.776 & 0.000 & 0.970 & 0.000 \\
othello-comp2007 & 1.000 & 1.000 & 0.988 & 0.950 & 0.285 & 0.250 & 0.470 & 0.000 \\
othellosuicide & 0.988 & 0.950 & 1.000 & 1.000 & 0.520 & 0.500 & 0.575 & 0.000 \\
pacman3p & 1.000 & 1.000 & 0.962 & 0.850 & 0.419 & 0.200 & 0.000 & 0.000 \\
pawnWhopping & 1.000 & 1.000 & 1.000 & 1.000 & 0.967 & 0.800 & 0.865 & 0.000 \\
platformJumpers & 0.949 & 0.900 & 0.647 & 0.550 & 0.713 & 0.350 & 0.000 & 0.000 \\
qyshinsu & 0.845 & 0.750 & 0.829 & 0.050 & 0.870 & 0.050 & 0.013 & 0.000 \\
rendezvous\_asteroids & 0.992 & 0.950 & 0.662 & 0.400 & 0.702 & 0.250 & 0.298 & 0.000 \\
rubikscube & 0.998 & 0.950 & 0.826 & 0.500 & 0.737 & 0.350 & 0.266 & 0.000 \\
snakeAssemblit & 0.906 & 0.700 & 0.188 & 0.100 & 0.247 & 0.050 & 0.000 & 0.000 \\
snake\_2009\_big & 0.946 & 0.700 & 0.869 & 0.700 & 0.681 & 0.050 & 0.857 & 0.000 \\
ticTacToeLarge & 1.000 & 1.000 & 0.844 & 0.750 & 0.745 & 0.650 & 0.486 & 0.000 \\
ticTacToeLargeSuicide & 1.000 & 1.000 & 0.403 & 0.050 & 0.692 & 0.550 & 0.486 & 0.000 \\
wallmaze & 1.000 & 1.000 & 0.948 & 0.800 & 0.918 & 0.600 & 0.971 & 0.000 \\
\midrule
\textbf{Average} & \textbf{0.915} & \textbf{0.837} & \textbf{0.852} & \textbf{0.661} & \textbf{0.672} & \textbf{0.407} & \textbf{0.400} & \textbf{0.014} \\
\bottomrule
\end{tabular}
\end{table*}

\begin{table*}[ht]
\centering
\small
\caption{Multistep state generation results with $n=3$ steps across all games.}
\label{tab:multistep_n3} 
\begin{tabular}{lcccccccc}
\toprule
 & \multicolumn{2}{c}{Gemini 2.5 Pro} & \multicolumn{2}{c}{Gemini 2.5 Flash} & \multicolumn{2}{c}{GPT-OSS 120B} & \multicolumn{2}{c}{Llama 3.3 70B} \\
\cmidrule(lr){2-3} \cmidrule(lr){4-5} \cmidrule(lr){6-7} \cmidrule(lr){8-9}
Game & JI & \%S & JI & \%S & JI & \%S & JI & \%S \\
\midrule
1reversi2 & 0.767 & 0.650 & 0.994 & 0.950 & 0.702 & 0.600 & 0.637 & 0.000 \\
battlebrushes & 0.200 & 0.200 & 0.894 & 0.650 & 0.182 & 0.000 & 0.508 & 0.000 \\
beatMania & 1.000 & 1.000 & 0.936 & 0.850 & 1.000 & 1.000 & 0.452 & 0.000 \\
bomberman2p & 1.000 & 1.000 & 0.999 & 0.950 & 0.324 & 0.000 & 0.009 & 0.000 \\
bomberman2p\_InvertedRoles & 1.000 & 1.000 & 0.999 & 0.950 & 0.324 & 0.000 & 0.009 & 0.000 \\
buttons & 1.000 & 1.000 & 1.000 & 1.000 & 0.980 & 0.900 & 0.693 & 0.300 \\
checkers & 1.000 & 1.000 & 0.873 & 0.750 & 0.476 & 0.100 & 0.000 & 0.000 \\
checkers-mustjump & 0.900 & 0.900 & 0.774 & 0.350 & 0.797 & 0.450 & 0.648 & 0.000 \\
checkers-newgoals & 0.752 & 0.750 & 0.893 & 0.600 & 0.763 & 0.300 & 0.532 & 0.000 \\
checkersSmall & 0.508 & 0.050 & 0.503 & 0.000 & 0.625 & 0.250 & 0.337 & 0.000 \\
checkersTiny & 0.306 & 0.000 & 0.522 & 0.000 & 0.474 & 0.050 & 0.335 & 0.000 \\
chess & 0.943 & 0.550 & 0.935 & 0.100 & 0.354 & 0.100 & 0.000 & 0.000 \\
chineseCheckers3 & 0.950 & 0.950 & 0.992 & 0.900 & 0.957 & 0.400 & 0.912 & 0.000 \\
cittaceot & 0.921 & 0.900 & 0.981 & 0.950 & 0.963 & 0.900 & 0.636 & 0.000 \\
connectfour & 1.000 & 1.000 & 1.000 & 1.000 & 0.651 & 0.600 & 0.922 & 0.900 \\
connectFourSuicide & 1.000 & 1.000 & 1.000 & 1.000 & 0.558 & 0.500 & 1.000 & 1.000 \\
dotsAndBoxes & 1.000 & 1.000 & 1.000 & 1.000 & 1.000 & 1.000 & 0.786 & 0.250 \\
dotsAndBoxesSuicide & 1.000 & 1.000 & 1.000 & 1.000 & 0.948 & 0.850 & 0.714 & 0.000 \\
farmers & 0.844 & 0.400 & 0.646 & 0.200 & 0.565 & 0.050 & 0.227 & 0.000 \\
fighter & 0.898 & 0.600 & 0.921 & 0.650 & 0.245 & 0.000 & 0.529 & 0.000 \\
god & 0.816 & 0.600 & 0.751 & 0.300 & 0.832 & 0.650 & 0.359 & 0.000 \\
mummymaze2p & 1.000 & 1.000 & 0.915 & 0.750 & 0.770 & 0.050 & 0.968 & 0.000 \\
othello-comp2007 & 0.978 & 0.900 & 0.895 & 0.750 & 0.693 & 0.600 & 0.598 & 0.000 \\
othellosuicide & 0.978 & 0.900 & 0.843 & 0.550 & 0.842 & 0.600 & 0.661 & 0.000 \\
pacman3p & 1.000 & 1.000 & 0.990 & 0.600 & 0.311 & 0.050 & 0.000 & 0.000 \\
pawnWhopping & 0.950 & 0.950 & 1.000 & 1.000 & 0.989 & 0.900 & 0.952 & 0.600 \\
platformJumpers & 0.898 & 0.800 & 0.848 & 0.700 & 0.601 & 0.250 & 0.000 & 0.000 \\
qyshinsu & 0.983 & 0.950 & 0.773 & 0.100 & 0.854 & 0.150 & 0.000 & 0.000 \\
rendezvous\_asteroids & 1.000 & 1.000 & 0.742 & 0.500 & 0.506 & 0.100 & 0.248 & 0.000 \\
rubikscube & 0.993 & 0.850 & 0.735 & 0.200 & 0.717 & 0.050 & 0.171 & 0.000 \\
snakeAssemblit & 0.865 & 0.550 & 0.353 & 0.100 & 0.186 & 0.050 & 0.000 & 0.000 \\
snake\_2009\_big & 0.896 & 0.550 & 0.891 & 0.550 & 0.654 & 0.000 & 0.818 & 0.000 \\
ticTacToeLarge & 1.000 & 1.000 & 0.972 & 0.950 & 0.756 & 0.650 & 0.472 & 0.050 \\
ticTacToeLargeSuicide & 1.000 & 1.000 & 0.714 & 0.500 & 0.674 & 0.550 & 0.442 & 0.000 \\
wallmaze & 1.000 & 1.000 & 0.997 & 0.800 & 0.915 & 0.500 & 0.954 & 0.000 \\
\midrule
\textbf{Average} & \textbf{0.896} & \textbf{0.800} & \textbf{0.865} & \textbf{0.634} & \textbf{0.662} & \textbf{0.377} & \textbf{0.472} & \textbf{0.089} \\
\bottomrule
\end{tabular}
\end{table*}

\begin{table*}[ht]
\centering
\small
\caption{Multistep state generation results with $n=4$ steps across all games.}
\label{tab:multistep_n4} 
\begin{tabular}{lcccccccc}
\toprule
 & \multicolumn{2}{c}{Gemini 2.5 Pro} & \multicolumn{2}{c}{Gemini 2.5 Flash} & \multicolumn{2}{c}{GPT-OSS 120B} & \multicolumn{2}{c}{Llama 3.3 70B} \\
\cmidrule(lr){2-3} \cmidrule(lr){4-5} \cmidrule(lr){6-7} \cmidrule(lr){8-9}
Game & JI & \%S & JI & \%S & JI & \%S & JI & \%S \\
\midrule
1reversi2 & 0.860 & 0.700 & 0.937 & 0.750 & 0.889 & 0.700 & 0.468 & 0.000 \\
battlebrushes & 0.600 & 0.600 & 0.926 & 0.850 & 0.238 & 0.000 & 0.447 & 0.000 \\
beatMania & 1.000 & 1.000 & 0.954 & 0.900 & 1.000 & 1.000 & 0.282 & 0.000 \\
bomberman2p & 0.938 & 0.850 & 0.899 & 0.400 & 0.358 & 0.050 & 0.052 & 0.000 \\
bomberman2p\_InvertedRoles & 0.938 & 0.850 & 0.899 & 0.400 & 0.358 & 0.050 & 0.052 & 0.000 \\
buttons & 1.000 & 1.000 & 1.000 & 1.000 & 0.966 & 0.900 & 0.627 & 0.200 \\
checkers & 1.000 & 1.000 & 0.938 & 0.600 & 0.584 & 0.250 & 0.000 & 0.000 \\
checkers-mustjump & 0.850 & 0.850 & 0.751 & 0.350 & 0.867 & 0.300 & 0.560 & 0.000 \\
checkers-newgoals & 0.800 & 0.750 & 0.984 & 0.700 & 0.896 & 0.350 & 0.490 & 0.000 \\
checkersSmall & 0.585 & 0.100 & 0.483 & 0.000 & 0.512 & 0.000 & 0.231 & 0.000 \\
checkersTiny & 0.194 & 0.000 & 0.571 & 0.000 & 0.475 & 0.000 & 0.271 & 0.000 \\
chess & 0.944 & 0.700 & 0.928 & 0.200 & 0.510 & 0.100 & 0.044 & 0.000 \\
chineseCheckers3 & 0.950 & 0.950 & 0.995 & 0.950 & 0.980 & 0.700 & 0.907 & 0.000 \\
cittaceot & 0.850 & 0.850 & 1.000 & 1.000 & 0.945 & 0.750 & 0.526 & 0.000 \\
connectfour & 1.000 & 1.000 & 1.000 & 1.000 & 0.821 & 0.800 & 0.667 & 0.000 \\
connectFourSuicide & 1.000 & 1.000 & 1.000 & 1.000 & 0.572 & 0.500 & 0.550 & 0.000 \\
dotsAndBoxes & 1.000 & 1.000 & 1.000 & 1.000 & 0.939 & 0.900 & 0.738 & 0.000 \\
dotsAndBoxesSuicide & 1.000 & 1.000 & 0.988 & 0.950 & 1.000 & 1.000 & 0.746 & 0.000 \\
farmers & 0.886 & 0.500 & 0.827 & 0.350 & 0.306 & 0.000 & 0.219 & 0.000 \\
fighter & 0.926 & 0.600 & 0.912 & 0.450 & 0.172 & 0.000 & 0.514 & 0.000 \\
god & 0.804 & 0.450 & 0.565 & 0.200 & 0.736 & 0.500 & 0.453 & 0.000 \\
mummymaze2p & 0.994 & 0.950 & 0.866 & 0.800 & 0.796 & 0.100 & 0.979 & 0.050 \\
othello-comp2007 & 0.930 & 0.850 & 0.868 & 0.650 & 0.613 & 0.550 & 0.475 & 0.000 \\
othellosuicide & 0.960 & 0.800 & 0.845 & 0.600 & 0.695 & 0.550 & 0.488 & 0.000 \\
pacman3p & 1.000 & 1.000 & 0.958 & 0.750 & 0.354 & 0.100 & 0.654 & 0.000 \\
pawnWhopping & 1.000 & 1.000 & 1.000 & 1.000 & 0.964 & 0.800 & 0.844 & 0.000 \\
platformJumpers & 0.900 & 0.900 & 0.648 & 0.500 & 0.590 & 0.050 & 0.000 & 0.000 \\
qyshinsu & 0.974 & 0.700 & 0.784 & 0.100 & 0.879 & 0.200 & 0.009 & 0.000 \\
rendezvous\_asteroids & 1.000 & 1.000 & 0.787 & 0.450 & 0.495 & 0.000 & 0.244 & 0.000 \\
rubikscube & 0.994 & 0.850 & 0.870 & 0.450 & 0.677 & 0.000 & 0.137 & 0.000 \\
snakeAssemblit & 0.895 & 0.350 & 0.224 & 0.050 & 0.236 & 0.000 & 0.000 & 0.000 \\
snake\_2009\_big & 0.895 & 0.500 & 0.937 & 0.350 & 0.677 & 0.000 & 0.850 & 0.000 \\
ticTacToeLarge & 0.850 & 0.850 & 1.000 & 1.000 & 0.844 & 0.650 & 0.667 & 0.000 \\
ticTacToeLargeSuicide & 1.000 & 1.000 & 0.800 & 0.650 & 0.951 & 0.900 & 0.621 & 0.000 \\
wallmaze & 1.000 & 1.000 & 0.982 & 0.650 & 0.930 & 0.500 & 0.956 & 0.000 \\
\midrule
\textbf{Average} & \textbf{0.901} & \textbf{0.786} & \textbf{0.861} & \textbf{0.601} & \textbf{0.681} & \textbf{0.379} & \textbf{0.451} & \textbf{0.007} \\
\bottomrule
\end{tabular}
\end{table*}

\begin{table*}[ht]
\centering
\small
\caption{Multistep state generation results with $n=5$ steps across all games.}
\label{tab:multistep_n5} 
\begin{tabular}{lcccccccc}
\toprule
 & \multicolumn{2}{c}{Gemini 2.5 Pro} & \multicolumn{2}{c}{Gemini 2.5 Flash} & \multicolumn{2}{c}{GPT-OSS 120B} & \multicolumn{2}{c}{Llama 3.3 70B} \\
\cmidrule(lr){2-3} \cmidrule(lr){4-5} \cmidrule(lr){6-7} \cmidrule(lr){8-9}
Game & JI & \%S & JI & \%S & JI & \%S & JI & \%S \\
\midrule
1reversi2 & 0.750 & 0.650 & 0.893 & 0.750 & 0.827 & 0.700 & 0.540 & 0.000 \\
battlebrushes & 0.300 & 0.300 & 1.000 & 1.000 & 0.197 & 0.000 & 0.427 & 0.000 \\
beatMania & 1.000 & 1.000 & 0.863 & 0.800 & 1.000 & 1.000 & 0.202 & 0.000 \\
bomberman2p & 1.000 & 1.000 & 0.936 & 0.450 & 0.407 & 0.050 & 0.020 & 0.000 \\
bomberman2p\_InvertedRoles & 1.000 & 1.000 & 0.936 & 0.450 & 0.407 & 0.050 & 0.020 & 0.000 \\
buttons & 1.000 & 1.000 & 0.967 & 0.950 & 0.980 & 0.900 & 0.624 & 0.250 \\
checkers & 0.791 & 0.650 & 0.816 & 0.550 & 0.639 & 0.400 & 0.000 & 0.000 \\
checkers-mustjump & 0.750 & 0.750 & 0.927 & 0.600 & 0.882 & 0.450 & 0.472 & 0.000 \\
checkers-newgoals & 0.900 & 0.850 & 0.986 & 0.550 & 0.853 & 0.150 & 0.466 & 0.000 \\
checkersSmall & 0.481 & 0.050 & 0.533 & 0.050 & 0.536 & 0.100 & 0.258 & 0.000 \\
checkersTiny & 0.359 & 0.000 & 0.540 & 0.000 & 0.484 & 0.050 & 0.293 & 0.000 \\
chess & 0.993 & 0.650 & 0.921 & 0.200 & 0.663 & 0.050 & 0.134 & 0.000 \\
chineseCheckers3 & 1.000 & 1.000 & 1.000 & 1.000 & 0.934 & 0.600 & 0.853 & 0.000 \\
cittaceot & 0.950 & 0.950 & 0.846 & 0.800 & 0.956 & 0.900 & 0.503 & 0.000 \\
connectfour & 1.000 & 1.000 & 1.000 & 1.000 & 0.708 & 0.600 & 0.954 & 0.950 \\
connectFourSuicide & 1.000 & 1.000 & 1.000 & 1.000 & 0.719 & 0.650 & 0.927 & 0.900 \\
dotsAndBoxes & 0.900 & 0.900 & 1.000 & 1.000 & 0.928 & 0.900 & 0.864 & 0.400 \\
dotsAndBoxesSuicide & 0.850 & 0.850 & 1.000 & 1.000 & 1.000 & 1.000 & 0.890 & 0.500 \\
farmers & 0.801 & 0.300 & 0.661 & 0.150 & 0.353 & 0.050 & 0.177 & 0.000 \\
fighter & 0.853 & 0.450 & 0.874 & 0.400 & 0.216 & 0.000 & 0.506 & 0.000 \\
god & 0.704 & 0.450 & 0.453 & 0.200 & 0.638 & 0.350 & 0.417 & 0.000 \\
mummymaze2p & 1.000 & 1.000 & 0.865 & 0.600 & 0.780 & 0.050 & 0.964 & 0.000 \\
othello-comp2007 & 0.906 & 0.750 & 0.836 & 0.450 & 0.711 & 0.650 & 0.613 & 0.000 \\
othellosuicide & 0.907 & 0.650 & 0.758 & 0.400 & 0.599 & 0.500 & 0.607 & 0.000 \\
pacman3p & 0.999 & 0.950 & 0.924 & 0.800 & 0.371 & 0.100 & 0.826 & 0.000 \\
pawnWhopping & 1.000 & 1.000 & 1.000 & 1.000 & 0.995 & 0.950 & 0.911 & 0.600 \\
platformJumpers & 0.848 & 0.800 & 0.793 & 0.400 & 0.618 & 0.150 & 0.000 & 0.000 \\
qyshinsu & 0.633 & 0.300 & 0.719 & 0.050 & 0.818 & 0.200 & 0.000 & 0.000 \\
rendezvous\_asteroids & 0.992 & 0.950 & 0.792 & 0.500 & 0.545 & 0.050 & 0.249 & 0.000 \\
rubikscube & 0.978 & 0.850 & 0.658 & 0.350 & 0.679 & 0.100 & 0.145 & 0.000 \\
snakeAssemblit & 0.744 & 0.150 & 0.270 & 0.100 & 0.135 & 0.000 & 0.000 & 0.000 \\
snake\_2009\_big & 0.894 & 0.500 & 0.897 & 0.650 & 0.828 & 0.050 & 0.855 & 0.000 \\
ticTacToeLarge & 1.000 & 1.000 & 0.950 & 0.900 & 0.844 & 0.650 & 0.804 & 0.550 \\
ticTacToeLargeSuicide & 1.000 & 1.000 & 0.905 & 0.800 & 0.955 & 0.800 & 0.773 & 0.500 \\
wallmaze & 1.000 & 1.000 & 0.997 & 0.800 & 0.876 & 0.400 & 0.951 & 0.000 \\
\midrule
\textbf{Average} & \textbf{0.865} & \textbf{0.734} & \textbf{0.843} & \textbf{0.591} & \textbf{0.688} & \textbf{0.389} & \textbf{0.493} & \textbf{0.133} \\
\bottomrule
\end{tabular}
\end{table*}

\begin{table*}[ht]
\centering
\small
\caption{Multistep state generation results with $n=7$ steps across all games.}
\label{tab:multistep_n7} 
\begin{tabular}{lcccccccc}
\toprule
 & \multicolumn{2}{c}{Gemini 2.5 Pro} & \multicolumn{2}{c}{Gemini 2.5 Flash} & \multicolumn{2}{c}{GPT-OSS 120B} & \multicolumn{2}{c}{Llama 3.3 70B} \\
\cmidrule(lr){2-3} \cmidrule(lr){4-5} \cmidrule(lr){6-7} \cmidrule(lr){8-9}
Game & JI & \%S & JI & \%S & JI & \%S & JI & \%S \\
\midrule
1reversi2 & 0.774 & 0.600 & 0.939 & 0.750 & 0.788 & 0.700 & 0.514 & 0.000 \\
battlebrushes & 0.550 & 0.550 & 1.000 & 1.000 & 0.271 & 0.000 & 0.516 & 0.000 \\
beatMania & 1.000 & 1.000 & 0.848 & 0.800 & 1.000 & 1.000 & 0.173 & 0.000 \\
bomberman2p & 0.939 & 0.750 & 0.914 & 0.250 & 0.336 & 0.000 & 0.175 & 0.000 \\
bomberman2p\_InvertedRoles & 0.939 & 0.750 & 0.914 & 0.250 & 0.336 & 0.000 & 0.175 & 0.000 \\
buttons & 1.000 & 1.000 & 1.000 & 1.000 & 0.937 & 0.850 & 0.614 & 0.250 \\
checkers & 0.997 & 0.950 & 0.905 & 0.650 & 0.725 & 0.500 & 0.000 & 0.000 \\
checkers-mustjump & 0.670 & 0.650 & 0.830 & 0.600 & 0.889 & 0.200 & 0.373 & 0.000 \\
checkers-newgoals & 0.802 & 0.800 & 0.939 & 0.550 & 0.869 & 0.200 & 0.420 & 0.000 \\
checkersSmall & 0.500 & 0.150 & 0.601 & 0.100 & 0.548 & 0.050 & 0.232 & 0.000 \\
checkersTiny & 0.331 & 0.000 & 0.551 & 0.050 & 0.527 & 0.000 & 0.223 & 0.000 \\
chess & 0.982 & 0.250 & 0.960 & 0.200 & 0.417 & 0.000 & 0.118 & 0.000 \\
chineseCheckers3 & 0.950 & 0.950 & 0.978 & 0.850 & 0.981 & 0.750 & 0.839 & 0.000 \\
cittaceot & 0.850 & 0.850 & 0.939 & 0.800 & 0.912 & 0.750 & 0.416 & 0.000 \\
connectfour & 0.950 & 0.950 & 1.000 & 1.000 & 0.866 & 0.750 & 0.749 & 0.600 \\
connectFourSuicide & 1.000 & 1.000 & 1.000 & 1.000 & 0.928 & 0.900 & 0.929 & 0.850 \\
dotsAndBoxes & 0.950 & 0.950 & 1.000 & 1.000 & 0.961 & 0.950 & 0.930 & 0.700 \\
dotsAndBoxesSuicide & 1.000 & 1.000 & 1.000 & 1.000 & 1.000 & 1.000 & 0.924 & 0.650 \\
farmers & 0.839 & 0.400 & 0.727 & 0.200 & 0.532 & 0.100 & 0.227 & 0.000 \\
fighter & 0.867 & 0.500 & 0.820 & 0.500 & 0.190 & 0.000 & 0.462 & 0.000 \\
god & 0.570 & 0.150 & 0.528 & 0.200 & 0.628 & 0.300 & 0.402 & 0.000 \\
mummymaze2p & 1.000 & 1.000 & 0.948 & 0.800 & 0.798 & 0.000 & 0.964 & 0.000 \\
othello-comp2007 & 0.947 & 0.700 & 0.762 & 0.300 & 0.458 & 0.350 & 0.559 & 0.050 \\
othellosuicide & 0.962 & 0.750 & 0.735 & 0.350 & 0.509 & 0.350 & 0.582 & 0.000 \\
pacman3p & 0.899 & 0.800 & 0.849 & 0.650 & 0.397 & 0.100 & 0.766 & 0.000 \\
pawnWhopping & 0.850 & 0.850 & 1.000 & 1.000 & 0.989 & 0.900 & 0.868 & 0.250 \\
platformJumpers & 0.796 & 0.500 & 0.643 & 0.300 & 0.596 & 0.100 & 0.000 & 0.000 \\
qyshinsu & 0.842 & 0.600 & 0.830 & 0.150 & 0.762 & 0.100 & 0.186 & 0.000 \\
rendezvous\_asteroids & 1.000 & 1.000 & 0.752 & 0.400 & 0.489 & 0.000 & 0.182 & 0.000 \\
rubikscube & 0.971 & 0.900 & 0.835 & 0.350 & 0.546 & 0.050 & 0.123 & 0.000 \\
snakeAssemblit & 0.791 & 0.100 & 0.197 & 0.000 & 0.196 & 0.000 & 0.000 & 0.000 \\
snake\_2009\_big & 0.538 & 0.250 & 0.892 & 0.350 & 0.692 & 0.000 & 0.841 & 0.000 \\
ticTacToeLarge & 0.950 & 0.950 & 0.977 & 0.950 & 0.904 & 0.650 & 0.904 & 0.750 \\
ticTacToeLargeSuicide & 1.000 & 1.000 & 1.000 & 1.000 & 0.922 & 0.850 & 0.769 & 0.400 \\
wallmaze & 1.000 & 1.000 & 0.892 & 0.550 & 0.893 & 0.500 & 0.944 & 0.000 \\
\midrule
\textbf{Average} & \textbf{0.857} & \textbf{0.703} & \textbf{0.849} & \textbf{0.569} & \textbf{0.680} & \textbf{0.370} & \textbf{0.489} & \textbf{0.129} \\
\bottomrule
\end{tabular}
\end{table*}

\begin{table*}[ht]
\centering
\small
\caption{Multistep state generation results with $n=10$ steps across all games.}
\label{tab:multistep_n10} 
\begin{tabular}{lcccccccc}
\toprule
 & \multicolumn{2}{c}{Gemini 2.5 Pro} & \multicolumn{2}{c}{Gemini 2.5 Flash} & \multicolumn{2}{c}{GPT-OSS 120B} & \multicolumn{2}{c}{Llama 3.3 70B} \\
\cmidrule(lr){2-3} \cmidrule(lr){4-5} \cmidrule(lr){6-7} \cmidrule(lr){8-9}
Game & JI & \%S & JI & \%S & JI & \%S & JI & \%S \\
\midrule
1reversi2 & 0.804 & 0.650 & 0.883 & 0.650 & 0.803 & 0.700 & 0.388 & 0.000 \\
battlebrushes & 0.900 & 0.900 & 0.943 & 0.800 & 0.266 & 0.000 & 0.385 & 0.000 \\
beatMania & 1.000 & 1.000 & 0.895 & 0.800 & 0.985 & 0.850 & 0.132 & 0.000 \\
bomberman2p & 0.986 & 0.900 & 0.854 & 0.100 & 0.336 & 0.000 & 0.221 & 0.000 \\
bomberman2p\_InvertedRoles & 0.986 & 0.900 & 0.854 & 0.100 & 0.336 & 0.000 & 0.221 & 0.000 \\
buttons & 1.000 & 1.000 & 1.000 & 1.000 & 0.943 & 0.900 & 0.578 & 0.250 \\
checkers & 0.794 & 0.600 & 0.844 & 0.500 & 0.619 & 0.200 & 0.000 & 0.000 \\
checkers-mustjump & 0.853 & 0.800 & 0.970 & 0.800 & 0.815 & 0.250 & 0.352 & 0.000 \\
checkers-newgoals & 0.801 & 0.750 & 0.932 & 0.600 & 0.869 & 0.250 & 0.376 & 0.000 \\
checkersSmall & 0.396 & 0.050 & 0.598 & 0.150 & 0.562 & 0.100 & 0.214 & 0.000 \\
checkersTiny & 0.550 & 0.050 & 0.588 & 0.100 & 0.430 & 0.000 & 0.214 & 0.000 \\
chess & 0.975 & 0.250 & 0.849 & 0.250 & 0.616 & 0.100 & 0.611 & 0.000 \\
chineseCheckers3 & 0.900 & 0.900 & 0.996 & 0.950 & 0.877 & 0.300 & 0.699 & 0.000 \\
cittaceot & 0.965 & 0.900 & 0.869 & 0.600 & 0.930 & 0.800 & 0.329 & 0.000 \\
connectfour & 0.850 & 0.850 & 0.983 & 0.900 & 0.764 & 0.600 & 0.513 & 0.000 \\
connectFourSuicide & 0.950 & 0.950 & 1.000 & 1.000 & 0.922 & 0.850 & 0.644 & 0.000 \\
dotsAndBoxes & 0.850 & 0.850 & 0.950 & 0.950 & 1.000 & 1.000 & 0.791 & 0.000 \\
dotsAndBoxesSuicide & 1.000 & 1.000 & 0.984 & 0.950 & 1.000 & 1.000 & 0.817 & 0.000 \\
farmers & 0.699 & 0.250 & 0.563 & 0.050 & 0.451 & 0.000 & 0.206 & 0.000 \\
fighter & 0.830 & 0.350 & 0.800 & 0.250 & 0.311 & 0.000 & 0.460 & 0.000 \\
god & 0.434 & 0.150 & 0.362 & 0.050 & 0.509 & 0.250 & 0.342 & 0.000 \\
mummymaze2p & 1.000 & 1.000 & 0.895 & 0.650 & 0.807 & 0.000 & 0.975 & 0.050 \\
othello-comp2007 & 0.890 & 0.500 & 0.760 & 0.350 & 0.544 & 0.450 & 0.453 & 0.000 \\
othellosuicide & 0.909 & 0.600 & 0.688 & 0.300 & 0.686 & 0.450 & 0.471 & 0.000 \\
pacman3p & 0.999 & 0.950 & 0.905 & 0.450 & 0.321 & 0.000 & 0.615 & 0.000 \\
pawnWhopping & 0.850 & 0.850 & 1.000 & 1.000 & 1.000 & 1.000 & 0.694 & 0.000 \\
platformJumpers & 0.948 & 0.850 & 0.742 & 0.300 & 0.600 & 0.050 & 0.000 & 0.000 \\
qyshinsu & 0.492 & 0.300 & 0.765 & 0.050 & 0.722 & 0.000 & 0.236 & 0.000 \\
rendezvous\_asteroids & 1.000 & 1.000 & 0.665 & 0.200 & 0.438 & 0.050 & 0.146 & 0.000 \\
rubikscube & 0.963 & 0.700 & 0.596 & 0.000 & 0.348 & 0.000 & 0.130 & 0.000 \\
snakeAssemblit & 0.787 & 0.200 & 0.247 & 0.050 & 0.223 & 0.000 & 0.000 & 0.000 \\
snake\_2009\_big & 0.896 & 0.650 & 0.888 & 0.350 & 0.693 & 0.150 & 0.787 & 0.000 \\
ticTacToeLarge & 1.000 & 1.000 & 0.996 & 0.950 & 0.898 & 0.700 & 0.833 & 0.000 \\
ticTacToeLargeSuicide & 0.950 & 0.950 & 0.953 & 0.850 & 0.962 & 0.850 & 0.833 & 0.000 \\
wallmaze & 1.000 & 1.000 & 0.946 & 0.700 & 0.914 & 0.450 & 0.942 & 0.000 \\
\midrule
\textbf{Average} & \textbf{0.863} & \textbf{0.703} & \textbf{0.822} & \textbf{0.507} & \textbf{0.671} & \textbf{0.351} & \textbf{0.446} & \textbf{0.009} \\
\bottomrule
\end{tabular}
\end{table*}


\subsection{Multistep action-state generation}

\noindent Table~\ref{tab:action_state} presents Multistep action--state generation results across all 35 games, extending the selected-game results shown in Table~4 in the main text.

\begin{table*}[ht]
\centering
\small
\caption{Multistep action-state generation results across all games.}
\label{tab:action_state}
\begin{tabular}{lcccccccc}
\toprule
 & \multicolumn{2}{c}{Gemini 2.5 Pro} & \multicolumn{2}{c}{Gemini 2.5 Flash} & \multicolumn{2}{c}{GPT-OSS 120B} & \multicolumn{2}{c}{Llama 3.3 70B} \\
\cmidrule(lr){2-3} \cmidrule(lr){4-5} \cmidrule(lr){6-7} \cmidrule(lr){8-9}
Game & JI & \%S & JI & \%S & JI & \%S & JI & \%S \\
\midrule
1reversi2 & 0.733 & 0.400 & 0.462 & 0.200 & 0.435 & 0.300 & 0.000 & 0.000 \\
battlebrushes & 0.500 & 0.500 & 0.868 & 0.850 & 0.213 & 0.000 & 0.087 & 0.000 \\
beatMania & 1.000 & 1.000 & 1.000 & 1.000 & 1.000 & 1.000 & 0.178 & 0.000 \\
bomberman2p & 0.744 & 0.700 & 0.896 & 0.650 & 0.387 & 0.000 & 0.091 & 0.000 \\
bomberman2p\_InvertedRoles & 0.744 & 0.700 & 0.896 & 0.650 & 0.387 & 0.000 & 0.091 & 0.000 \\
buttons & 0.980 & 0.950 & 1.000 & 1.000 & 0.988 & 0.950 & 0.143 & 0.000 \\
checkers & 0.948 & 0.850 & 0.494 & 0.400 & 0.412 & 0.250 & 0.000 & 0.000 \\
checkers-mustjump & 0.742 & 0.700 & 0.150 & 0.150 & 0.275 & 0.150 & 0.000 & 0.000 \\
checkers-newgoals & 0.997 & 0.950 & 0.499 & 0.450 & 0.443 & 0.050 & 0.000 & 0.000 \\
checkersSmall & 0.585 & 0.100 & 0.367 & 0.100 & 0.533 & 0.300 & 0.000 & 0.000 \\
checkersTiny & 0.443 & 0.000 & 0.158 & 0.050 & 0.429 & 0.250 & 0.000 & 0.000 \\
chess & 0.915 & 0.100 & 0.763 & 0.250 & 0.633 & 0.150 & 0.000 & 0.000 \\
chineseCheckers3 & 0.950 & 0.950 & 0.691 & 0.650 & 0.889 & 0.700 & 0.000 & 0.000 \\
cittaceot & 0.950 & 0.950 & 0.948 & 0.900 & 0.458 & 0.250 & 0.397 & 0.000 \\
connectfour & 1.000 & 1.000 & 0.961 & 0.950 & 0.763 & 0.700 & 0.380 & 0.000 \\
connectFourSuicide & 1.000 & 1.000 & 1.000 & 1.000 & 0.885 & 0.800 & 0.833 & 0.750 \\
dotsAndBoxes & 0.805 & 0.750 & 0.000 & 0.000 & 0.450 & 0.450 & 0.685 & 0.200 \\
dotsAndBoxesSuicide & 0.336 & 0.200 & 0.000 & 0.000 & 0.209 & 0.200 & 0.707 & 0.400 \\
farmers & 0.595 & 0.550 & 0.663 & 0.300 & 0.365 & 0.200 & 0.012 & 0.000 \\
fighter & 0.651 & 0.350 & 0.468 & 0.250 & 0.583 & 0.250 & 0.111 & 0.000 \\
god & 0.630 & 0.100 & 0.065 & 0.000 & 0.196 & 0.150 & 0.000 & 0.000 \\
mummymaze2p & 0.994 & 0.950 & 0.961 & 0.750 & 0.796 & 0.000 & 0.145 & 0.000 \\
othello-comp2007 & 0.715 & 0.350 & 0.399 & 0.050 & 0.566 & 0.300 & 0.000 & 0.000 \\
othellosuicide & 0.573 & 0.350 & 0.214 & 0.150 & 0.531 & 0.050 & 0.000 & 0.000 \\
pacman3p & 0.898 & 0.850 & 0.400 & 0.400 & 0.310 & 0.050 & 0.000 & 0.000 \\
pawnWhopping & 0.950 & 0.950 & 0.838 & 0.650 & 0.950 & 0.950 & 0.000 & 0.000 \\
platformJumpers & 0.900 & 0.900 & 0.199 & 0.150 & 0.466 & 0.150 & 0.000 & 0.000 \\
qyshinsu & 0.978 & 0.500 & 0.628 & 0.200 & 0.240 & 0.000 & 0.137 & 0.000 \\
rendezvous\_asteroids & 1.000 & 1.000 & 0.950 & 0.850 & 0.862 & 0.300 & 0.002 & 0.000 \\
rubikscube & 0.984 & 0.900 & 0.794 & 0.200 & 0.950 & 0.900 & 0.000 & 0.000 \\
snakeAssemblit & 0.135 & 0.050 & 0.040 & 0.000 & 0.043 & 0.000 & 0.000 & 0.000 \\
snake\_2009\_big & 0.963 & 0.550 & 0.895 & 0.400 & 0.271 & 0.050 & 0.000 & 0.000 \\
ticTacToeLarge & 1.000 & 1.000 & 0.977 & 0.950 & 0.940 & 0.800 & 1.000 & 1.000 \\
ticTacToeLargeSuicide & 0.946 & 0.900 & 0.932 & 0.850 & 0.956 & 0.800 & 1.000 & 1.000 \\
wallmaze & 0.997 & 0.800 & 1.000 & 1.000 & 0.823 & 0.700 & 0.138 & 0.000 \\
\midrule
\textbf{Average} & \textbf{0.808} & \textbf{0.653} & \textbf{0.616} & \textbf{0.469} & \textbf{0.561} & \textbf{0.347} & \textbf{0.175} & \textbf{0.096} \\
\bottomrule
\end{tabular}
\end{table*}


\subsection{Obfuscation}

Tables~\ref{tab:obf_placeholder} -~\ref{tab:obf_random} report Multistep state generation results for $n=5$ with the three considered obfuscation approaches: placeholder terms, dictionary words, and random strings, respective, corresponding to the obfuscation results summarized in Table~5 in the main text.


\subsection{Statistical Significance of Obfuscation Results}

To rigorously assess the impact of semantic obfuscation on reasoning performance, we conducted pairwise Wilcoxon signed-rank tests including all 35 games comparing the \textit{Multistep state generation} performance ($\%S$) across the different obfuscation variants. The significance level was set at $\alpha = 0.05$. Table~\ref{tab:obfuscation_stats} summarizes the results.

\begin{table*}[ht]
    \centering
    \small
    \caption{Wilcoxon signed-rank test results for obfuscation experiments (Orig. = no obfuscation, Rand. = random strings, Dict. = dictionary words, Place. = term placeholder). Significant results ($p < 0.05$) are marked in \textbf{bold}. }
    \label{tab:obfuscation_stats}
    \renewcommand{\arraystretch}{1.2}
    \begin{tabular}{llr}
        \toprule
        \textbf{Model} & \textbf{Comparison Pair} & \textbf{$p$-value} \\
        \midrule
        \multirow{6}{*}{\textbf{Gemini 2.5 Pro}} 
        & Orig. vs Rand. & $0.1112$ \\
        & Orig. vs Dict. & $\mathbf{0.0005}$ \\
        & Orig. vs Place. & $\mathbf{0.0032}$ \\
        & Rand. vs Dict. & $\mathbf{0.0369}$ \\
        & Rand. vs Place. & $0.1260$ \\
        & Dict. vs Place. & $0.1980$ \\
        \midrule
        \multirow{6}{*}{\textbf{Gemini 2.5 Flash}} 
        & Orig. vs Rand. & $\mathbf{0.0001}$ \\
        & Orig. vs Dict. & $\mathbf{<0.0001}$ \\
        & Orig. vs Place. & $\mathbf{0.0005}$ \\
        & Rand. vs Dict. & $\mathbf{0.0033}$ \\
        & Rand. vs Place. & $0.6484$ \\
        & Dict. vs Place. & $\mathbf{0.0033}$ \\
        \midrule
        \multirow{6}{*}{\textbf{GPT-OSS 120B}} 
        & Orig. vs Rand. & $\mathbf{<0.0001}$ \\
        & Orig. vs Dict. & $\mathbf{<0.0001}$ \\
        & Orig. vs Place. & $\mathbf{<0.0001}$ \\
        & Rand. vs Dict. & $\mathbf{0.0413}$ \\
        & Rand. vs Place. & $0.5053$ \\
        & Dict. vs Place. & $\mathbf{0.0090}$ \\
        \midrule
        \multirow{6}{*}{\textbf{Llama 3.3 70B}} 
        & Orig. vs Rand. & $0.4099$ \\
        & Orig. vs Dict. & $0.0691$ \\
        & Orig. vs Place. & $0.0691$ \\
        & Rand. vs Dict. & $0.4099$ \\
        & Rand. vs Place. & $0.4099$ \\
        & Dict. vs Place. & $0.4099$ \\
        \bottomrule
    \end{tabular}
\end{table*}

For Gemini 2.5 Pro, the performance difference between the original game definitions and the `Random Strings' obfuscation is not statistically significant ($p=0.1112$). It suggests that the model effectively utilizes the structural logic of the rules and is less dependent on meaningful variable names compared to other models.

Both Gemini 2.5 Flash and GPT-OSS 120B exhibit statistically significant degradation ($p < 0.001$) across all obfuscated variants compared to the original setting. Furthermore, for these models, the `Dictionary Words' variant (using unrelated nouns) proved significantly more detrimental than `Random Strings' ($p < 0.05$), indicating that misleading semantics disrupt reasoning more than the absence of semantics.
    
For Llama 3.3 70B, no statistically significant differences were observed across any comparisons ($p > 0.05$) because the model's performance is limited by its fundamental reasoning capability on the task itself, rather than by the presence or absence of semantic cues.

\begin{table*}[ht]
\centering
\small
\caption{Obfuscated game reasoning results for Multistep state generation experiment with $n=5$ moves across all games. \textit{Term placeholder} obfuscation.}
\label{tab:obf_placeholder}
\begin{tabular}{lcccccccc}
\toprule
 & \multicolumn{2}{c}{Gemini 2.5 Pro} & \multicolumn{2}{c}{Gemini 2.5 Flash} & \multicolumn{2}{c}{GPT-OSS 120B} & \multicolumn{2}{c}{Llama 3.3 70B} \\
\cmidrule(lr){2-3} \cmidrule(lr){4-5} \cmidrule(lr){6-7} \cmidrule(lr){8-9}
Game & JI & \%S & JI & \%S & JI & \%S & JI & \%S \\
\midrule
1reversi2 & 0.875 & 0.700 & 0.842 & 0.550 & 0.069 & 0.000 & 0.036 & 0.000 \\
battlebrushes & 0.169 & 0.150 & 0.465 & 0.250 & 0.127 & 0.000 & 0.069 & 0.000 \\
beatMania & 0.843 & 0.500 & 0.936 & 0.800 & 0.736 & 0.350 & 0.069 & 0.000 \\
bomberman2p & 0.970 & 0.650 & 0.889 & 0.250 & 0.446 & 0.050 & 0.647 & 0.000 \\
bomberman2p\_InvertedRoles & 0.970 & 0.650 & 0.889 & 0.250 & 0.446 & 0.050 & 0.647 & 0.000 \\
buttons & 0.933 & 0.900 & 0.850 & 0.850 & 0.944 & 0.800 & 0.365 & 0.000 \\
checkers & 0.996 & 0.900 & 0.949 & 0.350 & 0.048 & 0.000 & 0.447 & 0.000 \\
checkers-mustjump & 0.965 & 0.550 & 0.882 & 0.150 & 0.081 & 0.000 & 0.169 & 0.000 \\
checkers-newgoals & 0.823 & 0.550 & 0.972 & 0.600 & 0.074 & 0.000 & 0.243 & 0.000 \\
checkersSmall & 0.506 & 0.000 & 0.507 & 0.000 & 0.085 & 0.000 & 0.182 & 0.000 \\
checkersTiny & 0.490 & 0.000 & 0.470 & 0.000 & 0.150 & 0.000 & 0.161 & 0.000 \\
chess & 0.941 & 0.550 & 0.904 & 0.150 & 0.192 & 0.000 & 0.347 & 0.000 \\
chineseCheckers3 & 0.999 & 0.950 & 1.000 & 1.000 & 0.821 & 0.500 & 0.625 & 0.000 \\
cittaceot & 0.569 & 0.200 & 0.544 & 0.100 & 0.401 & 0.000 & 0.084 & 0.000 \\
connectfour & 1.000 & 1.000 & 0.942 & 0.900 & 0.270 & 0.000 & 0.179 & 0.050 \\
connectFourSuicide & 1.000 & 1.000 & 0.900 & 0.900 & 0.210 & 0.100 & 0.136 & 0.000 \\
dotsAndBoxes & 1.000 & 1.000 & 0.960 & 0.800 & 0.247 & 0.100 & 0.462 & 0.000 \\
dotsAndBoxesSuicide & 1.000 & 1.000 & 1.000 & 1.000 & 0.274 & 0.150 & 0.492 & 0.000 \\
farmers & 0.526 & 0.300 & 0.566 & 0.200 & 0.015 & 0.000 & 0.084 & 0.000 \\
fighter & 0.818 & 0.350 & 0.549 & 0.100 & 0.076 & 0.000 & 0.182 & 0.000 \\
god & 0.487 & 0.150 & 0.392 & 0.100 & 0.247 & 0.000 & 0.334 & 0.000 \\
mummymaze2p & 0.995 & 0.600 & 0.972 & 0.500 & 0.607 & 0.000 & 0.746 & 0.000 \\
othello-comp2007 & 0.914 & 0.650 & 0.623 & 0.050 & 0.200 & 0.000 & 0.395 & 0.000 \\
othellosuicide & 0.900 & 0.650 & 0.714 & 0.250 & 0.130 & 0.000 & 0.376 & 0.000 \\
pacman3p & 0.947 & 0.750 & 0.956 & 0.300 & 0.231 & 0.000 & 0.872 & 0.000 \\
pawnWhopping & 0.952 & 0.950 & 0.934 & 0.800 & 0.735 & 0.600 & 0.257 & 0.000 \\
platformJumpers & 0.525 & 0.150 & 0.050 & 0.050 & 0.450 & 0.000 & 0.000 & 0.000 \\
qyshinsu & 0.687 & 0.000 & 0.741 & 0.050 & 0.257 & 0.000 & 0.026 & 0.000 \\
rendezvous\_asteroids & 0.992 & 0.950 & 0.799 & 0.300 & 0.526 & 0.050 & 0.105 & 0.000 \\
rubikscube & 0.914 & 0.500 & 0.915 & 0.600 & 0.131 & 0.000 & 0.143 & 0.000 \\
snakeAssemblit & 0.258 & 0.000 & 0.437 & 0.000 & 0.091 & 0.000 & 0.000 & 0.000 \\
snake\_2009\_big & 0.864 & 0.200 & 0.892 & 0.100 & 0.110 & 0.000 & 0.019 & 0.000 \\
ticTacToeLarge & 1.000 & 1.000 & 0.973 & 0.950 & 0.386 & 0.100 & 0.549 & 0.050 \\
ticTacToeLargeSuicide & 1.000 & 1.000 & 1.000 & 1.000 & 0.318 & 0.050 & 0.659 & 0.250 \\
wallmaze & 0.951 & 0.950 & 0.991 & 0.700 & 0.287 & 0.000 & 0.118 & 0.000 \\
\midrule
\textbf{Average} & \textbf{0.822} & \textbf{0.583} & \textbf{0.783} & \textbf{0.427} & \textbf{0.298} & \textbf{0.083} & \textbf{0.292} & \textbf{0.010} \\
\bottomrule
\end{tabular}
\end{table*}

\begin{table*}[ht]
\centering
\small
\caption{Obfuscated game reasoning results for Multistep state generation experiment with $n=5$ moves across all games. \textit{Dictionary words} obfuscation.}
\label{tab:obf_dictionary}
\begin{tabular}{lcccccccc}
\toprule
 & \multicolumn{2}{c}{Gemini 2.5 Pro} & \multicolumn{2}{c}{Gemini 2.5 Flash} & \multicolumn{2}{c}{GPT-OSS 120B} & \multicolumn{2}{c}{Llama 3.3 70B} \\
\cmidrule(lr){2-3} \cmidrule(lr){4-5} \cmidrule(lr){6-7} \cmidrule(lr){8-9}
Game & JI & \%S & JI & \%S & JI & \%S & JI & \%S \\
\midrule
1reversi2 & 0.606 & 0.000 & 0.483 & 0.050 & 0.070 & 0.000 & 0.019 & 0.000 \\
battlebrushes & 0.760 & 0.600 & 0.382 & 0.050 & 0.075 & 0.000 & 0.027 & 0.000 \\
beatMania & 0.850 & 0.550 & 0.945 & 0.800 & 0.220 & 0.000 & 0.043 & 0.000 \\
bomberman2p & 0.947 & 0.800 & 0.946 & 0.400 & 0.322 & 0.000 & 0.075 & 0.000 \\
bomberman2p\_InvertedRoles & 0.947 & 0.800 & 0.946 & 0.400 & 0.322 & 0.000 & 0.075 & 0.000 \\
buttons & 0.967 & 0.950 & 0.949 & 0.850 & 0.338 & 0.050 & 0.265 & 0.000 \\
checkers & 0.979 & 0.500 & 0.865 & 0.100 & 0.146 & 0.000 & 0.004 & 0.000 \\
checkers-mustjump & 0.962 & 0.400 & 0.825 & 0.050 & 0.309 & 0.000 & 0.000 & 0.000 \\
checkers-newgoals & 0.978 & 0.600 & 0.862 & 0.200 & 0.326 & 0.000 & 0.132 & 0.000 \\
checkersSmall & 0.556 & 0.050 & 0.451 & 0.000 & 0.140 & 0.000 & 0.050 & 0.000 \\
checkersTiny & 0.535 & 0.000 & 0.476 & 0.000 & 0.118 & 0.000 & 0.108 & 0.000 \\
chess & 0.966 & 0.250 & 0.952 & 0.200 & 0.100 & 0.000 & 0.502 & 0.000 \\
chineseCheckers3 & 1.000 & 1.000 & 1.000 & 1.000 & 0.664 & 0.300 & 0.697 & 0.000 \\
cittaceot & 0.936 & 0.800 & 0.777 & 0.400 & 0.282 & 0.000 & 0.572 & 0.000 \\
connectfour & 0.985 & 0.900 & 0.844 & 0.450 & 0.267 & 0.050 & 0.085 & 0.000 \\
connectFourSuicide & 0.952 & 0.900 & 0.843 & 0.500 & 0.174 & 0.050 & 0.094 & 0.000 \\
dotsAndBoxes & 0.989 & 0.950 & 0.852 & 0.600 & 0.174 & 0.000 & 0.369 & 0.000 \\
dotsAndBoxesSuicide & 1.000 & 1.000 & 0.879 & 0.500 & 0.233 & 0.050 & 0.377 & 0.000 \\
farmers & 0.431 & 0.150 & 0.425 & 0.000 & 0.101 & 0.000 & 0.057 & 0.000 \\
fighter & 0.798 & 0.300 & 0.647 & 0.150 & 0.036 & 0.000 & 0.154 & 0.000 \\
god & 0.504 & 0.200 & 0.332 & 0.000 & 0.156 & 0.000 & 0.389 & 0.000 \\
mummymaze2p & 0.988 & 0.750 & 0.932 & 0.400 & 0.657 & 0.000 & 0.975 & 0.000 \\
othello-comp2007 & 0.637 & 0.050 & 0.552 & 0.000 & 0.090 & 0.000 & 0.360 & 0.000 \\
othellosuicide & 0.609 & 0.050 & 0.543 & 0.000 & 0.171 & 0.000 & 0.392 & 0.000 \\
pacman3p & 0.970 & 0.250 & 0.967 & 0.200 & 0.158 & 0.000 & 0.219 & 0.000 \\
pawnWhopping & 1.000 & 1.000 & 0.994 & 0.900 & 0.450 & 0.000 & 0.572 & 0.000 \\
platformJumpers & 0.789 & 0.050 & 0.920 & 0.050 & 0.449 & 0.000 & 0.026 & 0.000 \\
qyshinsu & 0.798 & 0.050 & 0.705 & 0.100 & 0.246 & 0.000 & 0.093 & 0.000 \\
rendezvous\_asteroids & 0.917 & 0.700 & 0.748 & 0.250 & 0.119 & 0.000 & 0.149 & 0.000 \\
rubikscube & 0.838 & 0.450 & 0.831 & 0.150 & 0.087 & 0.000 & 0.147 & 0.000 \\
snakeAssemblit & 0.670 & 0.000 & 0.534 & 0.050 & 0.065 & 0.000 & 0.021 & 0.000 \\
snake\_2009\_big & 0.975 & 0.450 & 0.905 & 0.050 & 0.139 & 0.000 & 0.548 & 0.000 \\
ticTacToeLarge & 1.000 & 1.000 & 0.864 & 0.700 & 0.270 & 0.000 & 0.458 & 0.000 \\
ticTacToeLargeSuicide & 1.000 & 1.000 & 0.932 & 0.850 & 0.248 & 0.000 & 0.464 & 0.050 \\
wallmaze & 0.994 & 0.800 & 0.975 & 0.550 & 0.567 & 0.000 & 0.647 & 0.000 \\
\midrule
\textbf{Average} & \textbf{0.852} & \textbf{0.523} & \textbf{0.774} & \textbf{0.313} & \textbf{0.237} & \textbf{0.014} & \textbf{0.262} & \textbf{0.001} \\
\bottomrule
\end{tabular}%
\end{table*}

\begin{table*}[ht]
\centering
\small
\caption{Obfuscated game reasoning results for Multistep state generation experiment with $n=5$ moves across all games. \textit{Random strings} obfuscation.}
\label{tab:obf_random}
\begin{tabular}{lcccccccc}
\toprule
 & \multicolumn{2}{c}{Gemini 2.5 Pro} & \multicolumn{2}{c}{Gemini 2.5 Flash} & \multicolumn{2}{c}{GPT-OSS 120B} & \multicolumn{2}{c}{Llama 3.3 70B} \\
\cmidrule(lr){2-3} \cmidrule(lr){4-5} \cmidrule(lr){6-7} \cmidrule(lr){8-9}
Game & JI & \%S & JI & \%S & JI & \%S & JI & \%S \\
\midrule
1reversi2 & 0.895 & 0.700 & 0.714 & 0.150 & 0.084 & 0.000 & 0.143 & 0.000 \\
battlebrushes & 0.451 & 0.200 & 0.456 & 0.200 & 0.124 & 0.000 & 0.058 & 0.000 \\
beatMania & 0.736 & 0.200 & 0.968 & 0.850 & 0.552 & 0.300 & 0.038 & 0.000 \\
bomberman2p & 0.987 & 0.750 & 0.860 & 0.100 & 0.352 & 0.000 & 0.859 & 0.000 \\
bomberman2p\_InvertedRoles & 0.987 & 0.750 & 0.860 & 0.100 & 0.352 & 0.000 & 0.859 & 0.000 \\
buttons & 1.000 & 1.000 & 0.980 & 0.900 & 0.847 & 0.650 & 0.396 & 0.000 \\
checkers & 0.997 & 0.950 & 0.930 & 0.250 & 0.122 & 0.000 & 0.000 & 0.000 \\
checkers-mustjump & 0.996 & 0.900 & 0.812 & 0.050 & 0.311 & 0.000 & 0.089 & 0.000 \\
checkers-newgoals & 0.999 & 0.950 & 0.943 & 0.300 & 0.227 & 0.000 & 0.011 & 0.000 \\
checkersSmall & 0.530 & 0.000 & 0.496 & 0.000 & 0.210 & 0.000 & 0.352 & 0.000 \\
checkersTiny & 0.542 & 0.000 & 0.463 & 0.000 & 0.247 & 0.000 & 0.296 & 0.000 \\
chess & 0.986 & 0.550 & 0.903 & 0.250 & 0.171 & 0.000 & 0.008 & 0.000 \\
chineseCheckers3 & 1.000 & 1.000 & 0.999 & 0.950 & 0.901 & 0.400 & 0.000 & 0.000 \\
cittaceot & 0.810 & 0.550 & 0.751 & 0.350 & 0.531 & 0.000 & 0.082 & 0.000 \\
connectfour & 1.000 & 1.000 & 0.983 & 0.900 & 0.192 & 0.050 & 0.223 & 0.000 \\
connectFourSuicide & 1.000 & 1.000 & 0.983 & 0.950 & 0.167 & 0.000 & 0.233 & 0.000 \\
dotsAndBoxes & 0.989 & 0.950 & 0.983 & 0.900 & 0.420 & 0.250 & 0.442 & 0.000 \\
dotsAndBoxesSuicide & 1.000 & 1.000 & 0.988 & 0.950 & 0.149 & 0.000 & 0.475 & 0.050 \\
farmers & 0.751 & 0.350 & 0.720 & 0.300 & 0.066 & 0.000 & 0.061 & 0.000 \\
fighter & 0.827 & 0.300 & 0.705 & 0.150 & 0.116 & 0.000 & 0.127 & 0.000 \\
god & 0.584 & 0.200 & 0.317 & 0.000 & 0.179 & 0.000 & 0.222 & 0.000 \\
mummymaze2p & 0.999 & 0.950 & 0.950 & 0.500 & 0.584 & 0.000 & 0.896 & 0.000 \\
othello-comp2007 & 0.921 & 0.650 & 0.620 & 0.100 & 0.289 & 0.000 & 0.417 & 0.000 \\
othellosuicide & 0.885 & 0.450 & 0.674 & 0.150 & 0.133 & 0.000 & 0.333 & 0.000 \\
pacman3p & 0.975 & 0.450 & 0.907 & 0.400 & 0.437 & 0.000 & 0.897 & 0.000 \\
pawnWhopping & 1.000 & 1.000 & 0.957 & 0.800 & 0.742 & 0.250 & 0.468 & 0.000 \\
platformJumpers & 0.960 & 0.550 & 0.951 & 0.250 & 0.691 & 0.000 & 0.000 & 0.000 \\
qyshinsu & 0.913 & 0.000 & 0.772 & 0.050 & 0.166 & 0.000 & 0.272 & 0.000 \\
rendezvous\_asteroids & 0.962 & 0.850 & 0.856 & 0.500 & 0.186 & 0.000 & 0.190 & 0.000 \\
rubikscube & 0.986 & 0.900 & 0.799 & 0.300 & 0.383 & 0.000 & 0.149 & 0.000 \\
snakeAssemblit & 0.833 & 0.300 & 0.620 & 0.050 & 0.128 & 0.000 & 0.012 & 0.000 \\
snake\_2009\_big & 0.915 & 0.100 & 0.937 & 0.250 & 0.251 & 0.000 & 0.496 & 0.000 \\
ticTacToeLarge & 1.000 & 1.000 & 0.977 & 0.950 & 0.678 & 0.400 & 1.000 & 1.000 \\
ticTacToeLargeSuicide & 1.000 & 1.000 & 0.955 & 0.900 & 0.636 & 0.450 & 1.000 & 1.000 \\
wallmaze & 1.000 & 1.000 & 0.985 & 0.600 & 0.442 & 0.000 & 0.650 & 0.000 \\
\midrule
\textbf{Average} & \textbf{0.898} & \textbf{0.643} & \textbf{0.822} & \textbf{0.411} & \textbf{0.345} & \textbf{0.079} & \textbf{0.336} & \textbf{0.059} \\
\bottomrule
\end{tabular}
\end{table*}

\section{Detailed Correlation Results}

Table~\ref{tab:correlation} reports correlation coefficients between individual game properties and model performance, measured using the $\%S$ metric averaged across all tasks. Correlations are computed independently for each model. These detailed results correspond to and extend the summary correlations presented in Table~6 in the main text.

\begin{table*}[h!]
\centering
\small
\caption{Correlations between game properties and model performance.}
\label{tab:correlation}
\begin{tabular}{lrrrr}
\toprule
Property & Gemini 2.5 Pro & Gemini 2.5 Flash & GPT-OSS 120B & Llama 3.3 70B \\
\midrule
Converging? & 0.45 & 0.53 & 0.46 & 0.71 \\
Total rules\_NEXT & -0.49 & -0.60 & -0.43 & -0.45 \\
Total constant facts\_NEXT & -0.47 & -0.53 & -0.49 & -0.44 \\
Rules 'H-Index'\_NEXT & -0.48 & -0.49 & -0.42 & -0.50 \\
Total conditions\_NEXT & -0.49 & -0.54 & -0.41 & -0.44 \\
Max conditions per rule\_NEXT & -0.48 & -0.49 & -0.42 & -0.39 \\
Max rule depth\_NEXT & -0.53 & -0.51 & -0.32 & -0.32 \\
Avg rule depth\_NEXT & -0.52  & -0.46 & -0.26 & -0.38 \\
Top level rules\_NEXT & -0.36 & -0.48 & -0.35 & -0.39 \\
Top level rules\_LEGAL & -0.40 & -0.48 & -0.38 & -0.29 \\
Conditions 'H-Index'\_NEXT & -0.38 & -0.39 & -0.33 & -0.39 \\
Max arguments per condition\_NEXT & -0.33 & -0.34 & -0.38 & -0.38 \\
Max conditions per rule\_LEGAL & -0.44 & -0.35  & -0.31 & -0.33 \\
Conditions 'H-Index'\_LEGAL & -0.38 & -0.40 & -0.31 & -0.34 \\
Total conditions\_LEGAL & -0.35 & -0.42 & -0.30 & -0.34 \\
Total rules\_LEGAL & -0.32 & -0.43 & -0.31 & -0.32 \\
"Connection game" elements? & 0.31 & 0.33 & 0.21 & 0.50 \\
Rules 'H-Index'\_LEGAL & -0.34 & -0.36 & -0.27 & -0.34 \\
Diff Heuristic & -0.36 & -0.35 & -0.12 & -0.47 \\
Negative conditions\_NEXT & -0.38 & -0.26 & -0.36 & -0.23 \\
Avg conditions per rule\_NEXT & -0.37 & -0.27 & -0.22 & -0.36 \\
Max arguments per condition\_LEGAL & -0.35 & -0.34 & -0.24 & -0.29 \\
Top level rule types\_LEGAL & -0.34 & -0.39 & -0.30 & -0.18 \\
Or conditions\_NEXT & -0.34 & -0.35 & -0.23 & -0.27 \\
No. of Players & -0.25 & -0.22 & -0.36 & -0.27 \\
Avg conditions per rule\_LEGAL & -0.36 & -0.27 & -0.22 & -0.25 \\
Offboard resource management? & -0.26 & -0.28 & -0.27 & -0.21 \\
Max rule depth\_LEGAL & -0.27 & -0.32 & -0.18 & -0.24 \\
Avg rule depth\_LEGAL & 0.14 & 0.23 & 0.32 & 0.30 \\
Avg arguments per condition\_LEGAL & -0.29 & -0.25 & -0.19 & -0.25 \\
Recurrent rules\_NEXT & -0.24 & -0.25 & -0.23 & -0.23 \\
Total constant factsTop level rule types\_LEGAL & -0.26 & -0.24 & -0.20 & -0.22 \\
Avg arguments per condition\_NEXT & -0.26 & -0.21 & -0.15 & -0.24 \\
Avg FactCount & 0.05 & -0.09 & -0.41 & -0.34 \\
Recurrent rules\_LEGAL & -0.13 & -0.23 & -0.16 & -0.17 \\
Or conditions\_LEGAL & -0.15 & -0.21 & -0.14 & -0.16 \\
Has truly simultaneous moves? & -0.06 & -0.12 & -0.12 & -0.26 \\
2D Board? & -0.13 & -0.08 & -0.22 & 0.10 \\
Negative conditions\_LEGAL & -0.03 & -0.12 & -0.07 & -0.06 \\
Point counting? & -0.06 & 0.01  & 0.11 & -0.07 \\
\bottomrule
\end{tabular}
\end{table*}

\begin{table*}[ht]
\centering
\small
\caption{Property values for all 35 games for top 10 highest correlated properties.}
\label{tab:all_games_features}
\begin{tabular}{lcccccccccc}
Game & \rotatebox{90}{Converging?} & \rotatebox{90}{Total rules\_NEXT} & \rotatebox{90}{Total constant facts\_NEXT} & \rotatebox{90}{Rules 'H-Index'\_NEXT} & \rotatebox{90}{Total conditions\_NEXT} & \rotatebox{90}{Max conditions per rule\_NEXT} & \rotatebox{90}{Max rule depth\_NEXT} & \rotatebox{90}{Avg rule depth\_LEGAL} & \rotatebox{90}{Top level rules\_NEXT} & \rotatebox{90}{Top level rules\_LEGAL} \\
\midrule
1reversi2 & 0 & 10 & 50 & 4 & 35 & 7 & 4 & 2.6 & 4 & 3 \\
battlebrushes & 0 & 27 & 44 & 5 & 93 & 6 & 2 & 0.7 & 14 & 4 \\
beatMania & 0 & 14 & 59 & 2 & 24 & 3 & 1 & 0.0 & 8 & 7 \\
bomberman2p & 0 & 16 & 137 & 3 & 32 & 4 & 1 & 1.7 & 11 & 2 \\
bomberman2p\_InvertedRoles & 0 & 0 & 0 & 0 & 0 & 0 & 0 & 0.0 & 0 & 0 \\
buttons & 1 & 19 & 6 & 2 & 38 & 2 & 0 & 1.0 & 19 & 3 \\
checkers & 0 & 43 & 272 & 5 & 120 & 7 & 3 & 2.1 & 32 & 10 \\
checkers-mustjump & 0 & 47 & 131 & 5 & 124 & 7 & 3 & 2.8 & 32 & 10 \\
checkers-newgoals & 0 & 47 & 171 & 5 & 124 & 7 & 3 & 2.1 & 32 & 10 \\
checkersSmall & 0 & 45 & 268 & 5 & 126 & 7 & 4 & 2.7 & 32 & 10 \\
checkersTiny & 0 & 45 & 264 & 5 & 126 & 7 & 4 & 2.7 & 32 & 10 \\
chess & 0 & 102 & 266 & 6 & 230 & 7 & 4 & 3.7 & 46 & 12 \\
chineseCheckers3 & 0 & 8 & 59 & 2 & 15 & 4 & 0 & 1.7 & 8 & 2 \\
cittaceot & 1 & 37 & 25 & 6 & 164 & 6 & 3 & 0.0 & 12 & 3 \\
connectfour & 1 & 7 & 21 & 2 & 14 & 4 & 2 & 1.3 & 5 & 4 \\
connectFourSuicide & 1 & 7 & 21 & 2 & 14 & 4 & 2 & 1.3 & 5 & 4 \\
dotsAndBoxes & 1 & 20 & 25 & 4 & 50 & 5 & 3 & 0.7 & 12 & 2 \\
dotsAndBoxesSuicide & 1 & 20 & 25 & 4 & 50 & 5 & 3 & 0.7 & 12 & 2 \\
farmers & 0 & 33 & 612 & 4 & 80 & 7 & 2 & 0.4 & 18 & 6 \\
fighter & 0 & 71 & 137 & 9 & 306 & 16 & 4 & 0.3 & 30 & 16 \\
god & 1 & 13 & 52 & 3 & 32 & 6 & 2 & 1.1 & 6 & 2 \\
mummymaze2p & 1 & 18 & 59 & 3 & 35 & 3 & 2 & 0.0 & 8 & 4 \\
othello-comp2007 & 0 & 19 & 27 & 4 & 46 & 4 & 3 & 4.1 & 6 & 3 \\
othellosuicide & 0 & 19 & 27 & 4 & 46 & 4 & 3 & 4.1 & 6 & 3 \\
pacman3p & 0 & 16 & 189 & 4 & 41 & 6 & 1 & 1.6 & 9 & 4 \\
pawnWhopping & 1 & 8 & 0 & 1 & 9 & 2 & 1 & 1.8 & 5 & 3 \\
platformJumpers & 1 & 13 & 10 & 2 & 18 & 3 & 1 & 3.3 & 11 & 4 \\
qyshinsu & 0 & 54 & 137 & 5 & 130 & 10 & 8 & 0.4 & 20 & 4 \\
rendezvous\_asteroids & 0 & 47 & 203 & 4 & 114 & 4 & 0 & 1.0 & 47 & 6 \\
rubikscube & 1 & 10 & 201 & 3 & 31 & 7 & 2 & 1.0 & 6 & 7 \\
snake\_2009\_big & 0 & 16 & 106 & 3 & 37 & 4 & 1 & 0.9 & 11 & 1 \\
snakeAssemblit & 0 & 56 & 76 & 4 & 140 & 6 & 3 & 0.5 & 22 & 13 \\
ticTacToeLarge & 1 & 4 & 2 & 1 & 5 & 2 & 0 & 0.7 & 4 & 4 \\
ticTacToeLargeSuicide & 1 & 4 & 2 & 1 & 5 & 2 & 0 & 0.7 & 4 & 4 \\
wallmaze & 1 & 17 & 29 & 3 & 34 & 3 & 1 & 0.0 & 15 & 6 \\
\bottomrule
\end{tabular}
\end{table*}

\end{document}